\documentclass[journal,twoside,web]{ieeecolor}
\usepackage{generic}
\usepackage{cite}
\usepackage{amsmath,amssymb,amsfonts}
\usepackage{algorithmic}
\usepackage{graphicx}
\usepackage{algorithm,algorithmic}
\usepackage{hyperref}
\hypersetup{hidelinks}
\usepackage{textcomp}
\def\BibTeX{{\rm B\kern-.05em{\sc i\kern-.025em b}\kern-.08em
    T\kern-.1667em\lower.7ex\hbox{E}\kern-.125emX}}
\usepackage{booktabs}
\usepackage{subcaption}
\usepackage{multirow}

\newcommand{\system}{DuaLK}

\begin{document}
\title{Bridging Stepwise Lab-Informed Pretraining and Knowledge-Guided Learning for Diagnostic Reasoning}

% \author{First A. Author, \IEEEmembership{Fellow, IEEE}, Second B. Author, and Third C. Author Jr., \IEEEmembership{Member, IEEE}
\author{Pengfei Hu, 
% \IEEEmembership{Member, IEEE}, 
Chang Lu, Fei Wang, 
% \IEEEmembership{Senior Member, IEEE}, 
and Yue Ning
% , \IEEEmembership{Member, IEEE}
% \thanks{Manuscript received 14 April 2025.}
\thanks{Pengfei Hu is with 
the Department of Computer Science, Stevens Institute of Technology, 
Hoboken, NJ 07030 USA (e-mail: phu9@stevens.edu).}
\thanks{Chang Lu was with 
the Department of Computer Science, Stevens Institute of Technology, 
Hoboken, NJ 07030 USA. 
He is now the independent researcher (e-mail: luchang.cs@gmail.com).}
\thanks{Fei Wang is with 
the Weill Cornell Medicine, Cornell University, Boulder, NY 
10065 USA (e-mail: few2001@med.cornell.edu).}
\thanks{Yue Ning is now with the Department of Computer Science, Stevens Institute of Technology, Hoboken, NJ 07030 USA (e-mail: yue.ning@stevens.edu).}
}

\maketitle
\begin{abstract}

Despite the growing use of Electronic Health Records (EHR) for AI-assisted diagnosis prediction, most data-driven models struggle to incorporate clinically meaningful medical knowledge.
They often rely on limited ontologies, lacking structured reasoning capabilities and comprehensive coverage.
This raises an important research question: Will medical knowledge improve predictive models to support stepwise clinical reasoning as performed by human doctors?
To address this problem, we propose \system{}, a dual-expertise framework that combines two complementary sources of information.
For external knowledge, we construct a Diagnosis Knowledge Graph (KG) that encodes both hierarchical and semantic relations enriched by large language models (LLM).
To align with patient data, we further introduce a lab-informed proxy task that guides the model to follow a clinically consistent, stepwise reasoning process based on lab test signals.
Experimental results on two public EHR datasets demonstrate that \system{} consistently outperforms existing baselines across four clinical prediction tasks.
These findings highlight the potential of combining structured medical knowledge with individual-level clinical signals to achieve more accurate and interpretable diagnostic predictions.
The source code is publicly available on \url{https://github.com/humphreyhuu/DuaLK}.

\end{abstract}


\section{Introduction}

The digitization of patient information such as EHR has transformed healthcare in terms of data storage, information retrieval, and computational pattern recognition.
EHR-based prediction models, including those for diagnosis~\cite{LuRCKN21}, heart failure~\cite{ChoiXLDFXD20}, and readmission prediction tasks~\cite{shickel2017deep}, leverage EHR data to forecast patient outcomes. 
To reduce bias risks of predictions based on data patterns rather than clinical understanding, it is essential to integrate medical domain knowledge from both individual health records and public medical knowledge into the predictive process.
(1) Individual knowledge can help model capture progression patterns (e.g. disease complications) of personal health status on EHR datasets~\cite{ShangMXS19, LuRN23, poulain2024graph};
(2) Public knowledge bases and clinical websites have been used to enhance the understanding of medical concepts~\cite{ChoiBSSS17, MaYXCZG18, jiang2023graphcare}, which reveals that even simple hierarchical structures of diseases can improve prediction performance.

Despite recent progress, performing diagnostic decision-making in a manner similar to physicians remains a challenging goal, with several challenges ahead: 
\begin{enumerate}
    \item \textbf{Underutilized public medical knowledge with limited quality.}
    While external knowledge beyond the training data has been widely utilized by previous approaches, these models often rely on simple ontologies (e.g., the hierarchy in ICD-9-CM~\cite{world1988international}), which fail to capture the comprehensive medical knowledge required for clinical prediction~\cite{ChoiBSSS17, ShangMXS19}.
    Textual and other unstructured features~\cite{LuRN23, xu2023seqcare, xu2024ram} have also been explored to improve predictive performance, but semantic ambiguity and irrelevant context remain challenges.
    \item \textbf{Lack of diagnostic reasoning when using EHR data.}
    Some studies~\cite{ShangMXS19, Rasmy0XTZ21, PrakashCRV21} take advantage of customized pretraining tasks to fully exploit EHR data, overlooking the causal and procedural nature of clinical decision-making.
    There are also some models using co-occurrence graphs to learn medical interactions (e.g., complications)~\cite{LuRCKN21, LuRN23, poulain2024graph}, but the relations they capture are simple associations, failing to reflect the multi-step diagnostic reasoning that physicians typically follow.
    \item \textbf{Inefficiency of integrating medical knowledge in EHR.} 
    Most models rely on either public knowledge or individual health data independently, as combining both may incur high computational costs during deployment~\cite{boll2024graph}. 
    However, models that rely on a single source of data fail to capture generalized patterns, leading to biased predictions and degraded robustness.
\end{enumerate}

To overcome these limitations, a stepwise and knowledge-informed framework is required to better align with real-world clinical decision-making~\cite{xu2023seqcare}. 
Clinicians typically begin by ordering lab tests informed by patients' symptoms and medical history, interpreting the results, and making a diagnostic decision~\cite{sutton2020overview}.
Inspired by this process, we introduce a lab-informed pretraining strategy, in which lab tests are treated as intermediate clinical signals. 
It enables the model to perform diagnostic reasoning in a stepwise manner rather than all at once.
To enhance the quality of medical knowledge, a diagnosis KG is constructed by selecting highly relevant clinical entities (e.g., diseases, drugs, phenotypes) and incorporating structured triples transformed from unstructured textual information.
To mitigate the high computational cost during deployment, we decouple the KG from the model’s prediction phase. This design allows the use of large-scale knowledge graphs without introducing additional inference-time overhead, thereby improving scalability and efficiency.

Therefore, we propose \system{}, a \textbf{Dual}-Expertise Synergy framework, that collaboratively enhances representation by both \textbf{L}ab-informed pretraining and \textbf{K}nowledge-guided learning for diagnostic reasoning.
% that collaboratively enhance hidden representation by both public and individual knowledge to enable accurate and clinically aligned diagnosis predictions.
A diagnosis-specific KG is constructed from multiple clinical databases by using large language models (LLMs) to capture hierarchical and semantic relationships among medical concepts. 
Next, A lab-informed learning module is designed to utilize individual health data through graph neural networks and a customized pretraining proxy task based on lab test data to rectify patient-level embeddings. 
In general, these two expertise are collaborated within an encoder-decoder architecture, enabling mutual refinement and enhancing clinical prediction accuracy.
The main contributions of this paper are as follows:

\begin{enumerate}
    \item We construct a diagnosis knowledge graph (KG) with a bi-hierarchical structure upon multiple medical knowledge sources to support predictive modeling. Experimental results on two public EHR datasets demonstrate that the proposed KG significantly enhances diagnostic prediction performance (e.g., F1-score and recall), while maintaining compatibility with various baseline models through a flexible and adaptable embedding strategy.
    \item To the best of our knowledge, this is the first work to leverage lab test data for pretraining, modeling stepwise diagnostic reasoning through lab assignment and abnormality detection. Furthermore, supplementary experiments reveal that incorporating laboratory information significantly improves predictive performance, highlighting the crucial role of lab tests in clinical diagnosis.
    \item We propose a diagnostic prediction framework explicitly designed to fuse both public and individual medical knowledge. Our approach consistently outperforms state-of-the-art baselines, demonstrating strong generalization and adaptability—even as data volume or diagnostic code complexity increases.
\end{enumerate}

\section{Related Work}
\label{sec:relatedwork}
% Deep learning models have been commonly applied in predictive healthcare to provide guidance for preventive care. 
% Both RNN-based \cite{ChoiBSKSS16, ChoiBSSS16, MaCZYSG17, BaiZEV18, MaZWRWTMGG20} and CNN-based models \cite{NguyenTWV17, MaGWZWRTGM20} have been developed to capture temporal and spatial dependencies for health risk predictions.
% With the advent of graph neural networks (GNNs), GRAM~\cite{ChoiBSSS17} and its successors incorporate graph learning to further enhance hidden representations for medical concepts~\cite{LiuLDTWCYZ20, wu2021leveraging, LuRN23}. 
% Meanwhile, previous work such as G-BERT~\cite{ShangMXS19} takes advantage of transformers and their variants to fully utilize EHR data (i.e. single-admission patients) through various proxy tasks~\cite{PrakashCRV21, Rasmy0XTZ21, poulain2024graph}.

\begin{figure}[t!]
    \centering
    \includegraphics[width=0.48\textwidth]{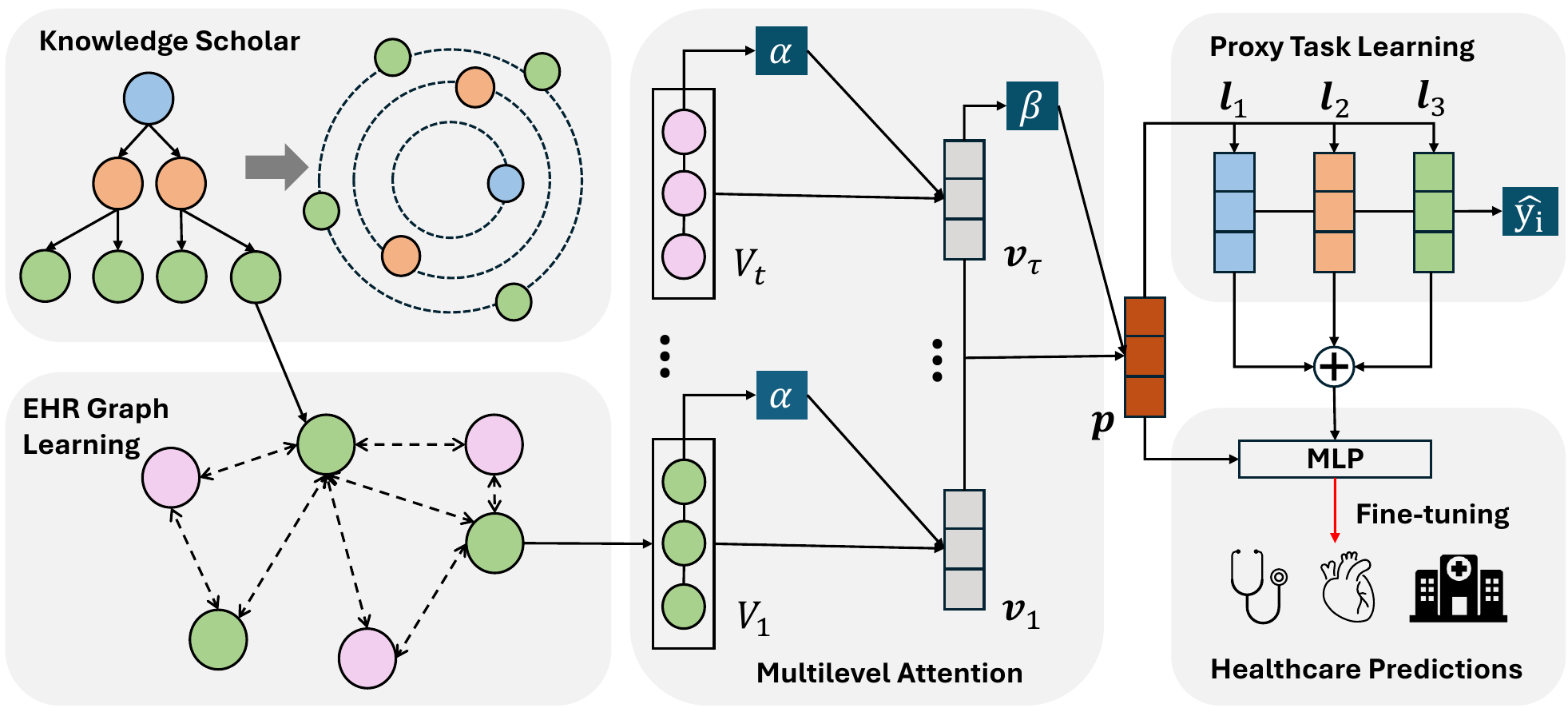}
    \caption{Overview of the proposed \system{} framework.}
    \label{fig:overview}
\end{figure}

\subsection{Clinical Prediction Models}
Deep learning models have been widely applied in predictive healthcare to provide guidance for preventive care. 
Both RNN-based \cite{ChoiBSKSS16, ChoiBSSS16, MaCZYSG17, BaiZEV18, MaZWRWTMGG20} and CNN-based models \cite{NguyenTWV17, MaGWZWRTGM20} have been developed to capture temporal and spatial dependencies for clinical risk predictions. 
With the rise of graph neural networks (GNNs), many works incorporate ontologies  over medical concepts (e.g. diseases) to enrich representations, such as CGL~\cite{LuRCKN21}, SeqCare~\cite{xu2023seqcare}, and GraphCare~\cite{jiang2023graphcare}.
Meanwhile, Transformer-based models like G-BERT~\cite{ShangMXS19} and its extensions leverage proxy tasks to extract information from structured EHR data, particularly for patients with limited visit histories~\cite{PrakashCRV21, Rasmy0XTZ21, poulain2024graph}.
Most of these methods rely on graphs with simple, single-type relations and demonstrate superiority across various EHR prediction tasks, which highlights the potential of enhancing prediction by more complex, multi-relational medical knowledge.

\subsection{Knowledge-Guided Clinical Prediction}

Predictive models that leverage graph and transformer architectures have achieved state-of-the-art performance in various clinical prediction tasks, highlighting their potential to advance predictive healthcare through the use of medical knowledge.
Such knowledge can be categorized into individual knowledge and public knowledge bases, as outlined below:

\begin{itemize}
    \item \textbf{Individual Knowledge:}
    Prior studies such as CGL~\cite{LuRCKN21} and Sherbet~\cite{LuRN23} construct EHR graphs based on interactions (e.g., disease complications) among medical codes in admissions, capturing hidden associations for clinical prediction.
    Transformer-based models like G-BERT~\cite{ShangMXS19} also adopt customized pretraining tasks to exploit extra supervision signals from EHR data, thereby improving convergence and enhancing clinical prediction~\cite{rasmy2021med, poulain2024graph}.
    However, existing methods either rely on oversimplified associations among medical concepts or employ proxy tasks disconnected from clinical reasoning, limiting their ability to model diagnostic decision processes.
    \item \textbf{Public Knowledge:} 
    External knowledge is derived from public data sources beyond the EHR dataset.
    Researchers have explored simple ontologies such as the ICD-9 hierarchy~\cite{world1988international}, retrieval-augmented generation (RAG) methods with unstructured medical descriptions like RAM-EHR~\cite{xu2024ram}, and large-scale ontologies for prediction, such as in GraphCare~\cite{jiang2023graphcare}.
    However, they often emphasize either hierarchical or semantic knowledge in isolation, which may introduce noisy or misaligned information and impair prediction performance.
\end{itemize}

While public and individual knowledge each provide complementary insights, most existing studies rely on only one of them~\cite{boll2024graph}. 
Effectively combining both hierarchical and semantic knowledge sources is essential for improving the robustness and generalizability of predictive healthcare models.

\subsection{Clinical Prediction with Multi-aspect Features}

In addition to leveraging medical-domain knowledge, combining both structured and unstructured features also enhance predictions.
MiME~\cite{ChoiXSS18} and GCT~\cite{ChoiXLDFXD20} use graph structures with lab results to augment representations, while CGL~\cite{LuRCKN21} and MedGTX~\cite{ParkBKKC22} leverage unstructured clinical notes to highlight the value of additional information. 
However, most approaches rely heavily on these multi-aspect features, limiting their applicability to data without those features. 
Our proposed framework integrates laboratory features as auxiliary data in graph construction and proxy tasks, enhancing predictions even when no desired feature is available.

\subsection{Reasoning in Clinical Prediction}
Many studies treat graph learning as a form of reasoning based on neighboring entities~\cite{LuRN23, xu2023seqcare, jiang2023graphcare}, which often reduces reasoning to static associations.
Reinforcement learning-based methods attempt to emulate clinical inquiry processes through multi-step interactions~\cite{zou2024ai}; however, their reliance on iterative decision-making leads to high training complexity and substantial data requirements, making them difficult to scale and less applicable in real-world clinical settings~\cite{smith2023bias}.
More recently, LLMs have been directly used as predictors to simulate clinician-like reasoning~\cite{nguyen2024carer, jiang2025reasoningenhanced}, a strategy which is prone to hallucinations in medical contexts and involves high computational costs.
Overall, most of them fall short in explicitly modeling the stepwise and causally structured nature of clinical reasoning, underscoring the need for more faithful and efficient reasoning frameworks.

\section{Methodology}
In this section, we propose \system{}, a dual-expertise framework designed for predictive healthcare that aims to provide comprehensive diagnoses based on public and individual knowledge. 
An overview of \system{} is shown in Figure~\ref{fig:overview}, and the framework operates in three main steps:

\textit{Step 1:} Build a diagnosis-specific KG enriched by hierarchies upon diseases and phenotypes via LLM, then provide initial embeddings as priors for diseases.

\textit{Step 2:} Generate a disease-lab EHR graph for graph learning, and customize a lab-informed proxy task to provide stepwise guidance for prediction. 

\textit{Step 3:} Get patient-level embeddings through the encoder-decoder structure, refining public knowledge by lab results for accurate clinical prediction.

\subsection{General Notation}

An EHR dataset $\mathcal{S}$ is a collection of temporal admission records of $N$ patients 
$\{\mathbf{p}_1,\mathbf{p}_2,...,\mathbf{p}_N\}$, and $\mathbf{p}_u$ denotes patient $u$.
Such patient can also be represented as a sequence of $T_u$ admission records 
$(V_{1}^{u}, V_{2}^{u},.., V_{T_u}^{u}) \in \mathbf{p}_u$ in chronological order.
We omit the patient index $u$ in the following sections and explain our framework using a single patient as an example for clarity.
We denote the entire set of medical concepts as \(\mathcal{C}=\{c_1, c_2,..., c_{|\mathcal{C}|}\}\) where $|\mathcal{C}|$ is the vocabulary size of medical concepts. 
Each admission, like $t$-th admission $V_t \in (V_{1}, V_{2},.., V_{T})$, contains a subset of $\mathcal{C}$.
Note that, We consider diseases and lab results as medical concepts in proposed framework, and we denote diseases $\{d_1, d_2,..., d_{|\mathcal{D}|}\}$ and lab results $\{l_1, l_2,..., l_{|\mathcal{L}|}\}$ within vocabulary $\mathcal{D}, \mathcal{L} \subset \mathcal{C}$.

\begin{figure}[t!]
    \centering
    \includegraphics[width=0.48\textwidth]{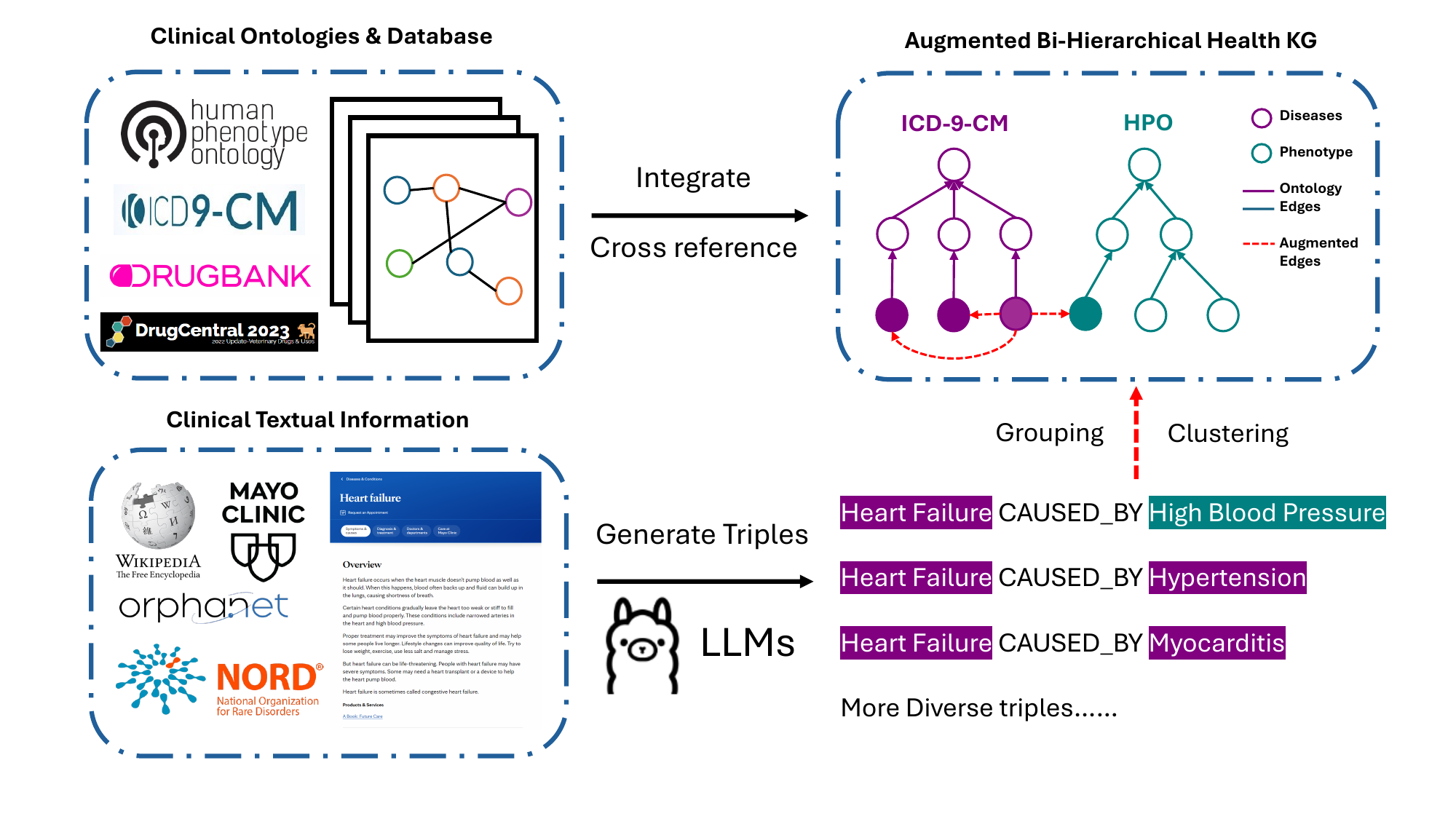}
    \caption{Overview of Augmented Bi-hierarchical KG.}
    \label{fig:kg_example}
\end{figure}

\subsection{Generate Diagnosis KG}

We establish knowledge scholar by converting both existing medical KGs and verified textual knowledge into the structured triplet set via the in-context-learning power of LLMs.
Figure~\ref{fig:kg_example} illustrates how to augment KG with textual information.
In general, we aim to integrate and standardize a uniform KG based on both medical ontologies and textual knowledge.

\subsubsection*{HKG Construction}
Healthcare KGs (HKG) integrate information from multiple datasets~\cite{cuietal2023}, providing a more complex structure than all-in-one ontologies such as UMLS~\cite{bodenreider2004unified}. 
We focus on constructing a healthcare KG that links diseases, drugs, and phenotypes to support precise clinical prediction. 
Therefore, five standardized ontologies and experimentally derived data are adopted for KG construction, ensuring broad coverage across key medical entities: 

\begin{enumerate}
    \item \textit{DrugBank}~\cite{knox2024drugbank}:
    The comprehensive resource containing detailed pharmaceutical knowledge. We retrieved the latest version (5.1.12), published on March 14, 2024. Our focus is on synergistic drug interactions, which represent the bidirectional connections between two drugs.
    \item \textit{DrugCentral}~\cite{ursu2016drugcentral}: 
    The verified database that provides drug-disease relations, including indications, contraindications, and off-label uses. We utilized the updated SQL Database, released on November 1, 2023, for our study.
    \item \textit{HPO}~\cite{gargano2024mode}: 
    The Human Phenotype Ontology offers details on phenotype abnormalities associated with diseases. We used the most recent updates, from April 19, 2024, focusing on disease-phenotype and phenotype-phenotype relationships, and extracted verified associations with expertly curated annotations.
    \item \textit{ICD-9-CM}~\cite{world1988international}: 
    The encoding system is to classify medical conditions for billing purposes, and it underlies the diagnostic information in most EHR data. For our purposes, we extracted the parent-child relationships among codes to describe disease interactions.
    \item \textit{SIDER}~\cite{kuhn2016sider}: 
    The Side Effect Resource provides side effects caused by various drugs. We retrieved this information from the data release dated October 21, 2015.
\end{enumerate}

The HPO and the ICD-9-CM hierarchies serve as the foundation for our KG, with bi-hierarchical structure enhancing its robustness and connectivity. 
We denote the resulting healthcare KG, sourced from existing databases, as $\mathcal{G}_{M}$, containing $M$ tuples of head, relation, and tail. 

\subsubsection*{LLM-KG Generation} 
We further extract relational triples from text information by leveraging the in-context learning capabilities of LLMs. 
Text information is parsed from the following four clinical websites/databases:

\begin{enumerate}
    \item \textit{Mayo Clinic}~\cite{brennan1998mayo} maintains a knowledge base with clinical information on over 2000 diseases.
    We collected Mayo Clinic web knowledge and mainly focus on both ``Symptoms \& Causes'' (Overview, Symptoms, When to see a doctor, Causes, Risk Factors, Complications, and Prevention) and ``Diagnosis \& Treatment'' parts for each condition.
    \item \textit{Orphanet}~\cite{weinreich2008orphanet} 
    is a clinical database dedicated to rare diseases. We collected data focusing on information about definitions, prevalence, treatment, and clinical descriptions for rare diseases.
    \item \textit{Rare Disease Database}~\cite{schieppati2008rare}
     contains supplementary knowledge source for uncovered rare diseases. We collected data to retrieve information on disease overviews, symptoms, causes, treatments, and clinical trials.
    \item \textit{Wikipedia} 
    is the most well-known online encyclopedia. 
    It designs a specific webpage for ICD-9-CM codes\footnote{The Wikipedia ICD-9-CM webpage: \url{https://en.wikipedia.org/wiki/List_of_ICD-9_codes}}. We retrieved related information for each disease on the separate webpage. For example, the page of ``Typhoid Fever'' provides detailed information on symptoms, causes, diagnosis, prevention, treatment, epidemiology, history, terminology, and societal impact, offering more comprehensive and diverse knowledge compared to other textual sources. 
\end{enumerate}

Incorporating unstructured textual information into the Diagnosis KG is achieved by leveraging the in-context learning capabilities of LLMs to extract relational triples from curated disease descriptions. As all inputs originate from verified or expert-curated sources, the risk of hallucination is minimal. A prompting-based approach is adopted instead of retrieval-augmented generation (RAG), showing better empirical performance—likely due to reduced semantic loss and the relatively low task complexity. 
The LLM-generated triples in the format \texttt{[ENTITY 1, RELATION, ENTITY 2]} are aggregated to form a knowledge graph $\mathcal{G}_{N}$ containing $N$ triples.

To further improve output robustness, two strategies are introduced during inference: 
% \yue{x and y are used before they are defined; can we remove or explain later when introduced?}
(1) \textbf{Iterated Prompting}, where the full input is re-prompted to ensure completeness; and (2) \textbf{Re-read Prompting}, where the same prompt is passed repeatedly for refinement. Final triples are manually reviewed and added to the Diagnosis KG as additional semantic edges, enriching disease node representations.
LLM-KG is performed using \texttt{Llama-3.1-8b}~\cite{grattafiori2024llama}, an open-source language model selected for its accessibility and compatibility with local deployment. 
However, other models like \texttt{DeepSeek}~\cite{guo2025deepseek}  are also compatible with the proposed workflow.
A detailed prompt example is provided in Appendix\ref{app:prompt_llm}.

\subsubsection*{Diagnosis-KG Integration}

To construct a unified Diagnosis KG $\mathcal{G}_{H}$, we integrate two components: the HKG $\mathcal{G}_M$ and the LLM-generated KG $\mathcal{G}_N$, as defined below: 

\begin{align}
\mathcal{G}_{M} &= \bigcup\nolimits_{M} ( 
\mathbf{h}^{(M)}_m, \mathbf{r}^{(M)}_m, \mathbf{t}^{(M)}_m ) \label{eq:gm} \\
\mathcal{G}_{N} &= \bigcup\nolimits_{x,y} \bigcup\nolimits_{N} ( 
\mathbf{h}^{(N)}_n, \mathbf{r}^{(N)}_n, \mathbf{t}^{(N)}_n ) \label{eq:gn} \\
\mathcal{G}_{H} &= \mathcal{G}_{M} \cup \mathcal{G}_{N}, \:
\tilde{\mathcal{G}_{H}} = \textsc{Normalize}(\mathcal{G}_{H}) \label{eq:gh}
\end{align}
Here, $x$ means the number of iteration for iterative running and and $y$ denotes the number of times the generated output is re-evaluated (re-reading) to refine the quality of triples.

\subsubsection*{KG Normalization} 
Our approach for normalizing the combined KG $\mathcal{G}_{H}$ involves two key steps:
\textbf{(1) Uniform encoding system for entities:} 
In addition to phenotype nodes sourced from HPO, we align disease and drug entities using the ICD-9-CM~\cite{world1988international} and ATC~\cite{nahler2009anatomical} coding systems, both of which are widely used in EHR data. 
A cross-referencing method is employed to convert $\mathcal{G}_{M}$ and $\mathcal{G}_{N}$ into a shared hierarchy. 
Given that drugs are not the primary focus, we use ATC-4 to simplify the numerous synergistic interactions in DrugBank.
\textbf{(2) Clustering duplicated or unmatched triples:} 
The LLM-prompting process is run multiple times, which can generate duplicated nodes and edges.
Moreover, cross-referencing files may not cover all concepts, and unmatched entities leave noise in KG. 
To address these problems, we calculate cosine similarity for nodes and edges based on their embeddings. 
Using a predefined threshold $\theta$, we group entities within categories (i.e. ``disease'', ``drug'', ``phenotype'', and ``other''), and apply hierarchical clustering to reorganize edges.

\setlength{\tabcolsep}{3pt}
\begin{table}[t]
    \centering
    \small
    \caption{Statistical Comparison among KGs.
    (* means results might fluctuate by different parameter settings.)}
    \label{tab:knowledge_graphs}
    \begin{tabular}{lrrr}
    \toprule
    \textbf{KG} & \#Nodes & \#EdgeTypes & \#Triples \\
    \midrule
    \multicolumn{4}{l}{\textbf{GraphCare: }[Disease, Procedure, Drug, Other]} \\
    \qquad - Ontology* & 10,805 & 54 & 81,073 \\
    \qquad - GPT-4 output* & 4,599 & 752 & 31,325 \\
    \qquad - Final KG* & 12,284 & 785 & 104,460 \\
    \midrule
    \multicolumn{4}{l}{\textbf{PrimeKG: }[Disease, Drug, Protein ... 10 in total]} \\
    \qquad - Public KG & \textbf{129,375} & 30 & \textbf{4,050,249} \\
    \qquad - Text Description & - & - & - \\
    \midrule
    \multicolumn{4}{l}{\textbf{Diagnosis KG: }[Disease, Drug, Phenotype, Other]} \\
    \qquad - HKG & 32,941 & 6 & 477,992 \\
    \qquad - LLM-KG* & 42,436 & \textbf{3,642} & 82,191 \\
    \qquad - Final KG* & 52,604 & \textbf{3,645} & 560,183 \\
    \bottomrule
    \end{tabular}
\end{table}

\subsubsection*{Advancement Analysis}
To demonstrate the advancements of our proposed Diagnosis KG, we compare it with two recent health-related KGs: GraphCare~\cite{jiang2023graphcare} and PrimeKG~\cite{chandak2023building}.
A statistical summary of these KGs is presented in Table~\ref{tab:knowledge_graphs}. 
The analysis highlights the following advantages of our KG:

\begin{enumerate}
    \item Compared to GraphCare~\cite{jiang2023graphcare}, our KG is larger in scale and is built upon the widely-used ICD-9-CM and HPO system, offering hierarchical and precise knowledge. 
    In contrast, GraphCare relies on zero-shot generation from LLMs without fine-tuning, which can result in incorrect triples due to hallucination~\cite{xu2024ram}, along with unrelated entities (e.g. social impact) that are irrelevant to the clinical task.
    Furthermore, the subgraph sampling process can introduce instability into the output.
    Hence, our KG provides more stable and accurate results in downstream prediction tasks.
    \item Compared to PrimeKG~\cite{chandak2023building}, our KG includes more edge types, primarily augmented by LLM-generated triples. The LLM reasoning process not only enhances connectivity within the KG but also converts unstructured text into structured triples, making information retrieval more efficient from single-modal data.
\end{enumerate}

Although PrimeKG covers a broader range of medical concepts, it is more prone to data quality issues that can introduce bias into EHR predictions.
For example, genes and proteins included in PrimeKG are seldom considered in physicians' decision-making processes, particularly when it comes to diagnosis.
In contrast, our diagnosis-driven KG contains numerous normalized triples with diverse edges, enhanced by LLMs' in-context learning capabilities, making it highly reproducible and well-suited for various diagnosis-related healthcare predictions.
Such analysis will then be supported by the results of the ablation study.

\subsection{Polar-space KG Embedding}
Traditional KG embedding methods often emphasize either contextual or hierarchical information, limiting their ability to represent both simultaneously. In our setting, the constructed Diagnosis KG encodes both types of signals, requiring an embedding method capable of capturing their interplay. To meet this need, we adopt polar-space embedding~\cite{ZhangCZW20}, which balances semantic expressiveness with computational efficiency and interpretability. A comparison with other KG embedding methods~\cite{BordesUGWY13, NickelK17, SunDNT19, FiondaP20} is provided in Appendix~\ref{app:kge_compare}.

To describe medical concepts from both contextual and hierarchical perspectives, we leverage two properties of the polar coordinate system:
(1) \textbf{Radial coordinate:} map entities across different levels of the hierarchy;
(2) \textbf{Angular coordinate:} map entities at the same level based on contextual information.
We denote entities, including head $\mathbf{h}$ and tail $\mathbf{t}$, as $\mathbf{e}$, and relation embeddings as $\mathbf{r}$.
Subscripts, such as $\mathbf{e}_{r}$ and $\mathbf{e}_{a}$, represent the modulus and phase components, respectively. 
Each entry of $\mathbf{h}_{r,a}$ or $\mathbf{t}_{r,a}$ corresponds to a radial or angular value, while each entry of $\mathbf{r}_{r,a}$ represents a scaling transformation between the corresponding head and tail.
The radial and angular coordinates can be formulated as follows:
% \[
% \left\{
% \begin{array}{l}
% \mathbf{h}_r \circ \mathbf{r}_r = \mathbf{t}_r, \text{where} \ \mathbf{h}_r, \mathbf{t}_r \in \mathbb{R}^k, \ \mathbf{r}_r \in \mathbb{R}_+^k, \\ [3pt]
% (\mathbf{h}_a + \mathbf{r}_a) \mod 2\pi = \mathbf{t}_a, \text{where} \ \mathbf{h}_a, \mathbf{t}_a, \mathbf{r}_a \in [0, 2\pi)^k,
% \\ [3pt]
% \mathbf{h} = [\mathbf{h}_r;\mathbf{h}_a], \text{where} (\mathbf{h}_{r,i}, \mathbf{a}_{a,i}) \text{ is a 2D point}.
% \end{array}
% \right.
% \]
\begin{gather}
\mathbf{h}_r \circ \mathbf{r}_r = \mathbf{t}_r, \:
 \ \mathbf{h}_r, \mathbf{t}_r, \mathbf{r}_r \in \mathbb{R}^k \label{eq:kg1} \\
(\mathbf{h}_a + \mathbf{r}_a) \bmod 2\pi = \mathbf{t}_a, \:
 \ \mathbf{h}_a, \mathbf{t}_a, \mathbf{r}_a \in [0, 2\pi)^k
\end{gather}
% From both radial and angular perspectives, we map entities into the polar coordinate system and calculate the corresponding circular arc distance by simple weighted sum:
According to the coordinates for triples, we can further calculate radial and angular distances as follows:
\begin{align}
    d_{r}(\mathbf{h}_r, \mathbf{t}_r) &= \|\mathbf{h}_r \circ \mathbf{r}_r - \mathbf{t}_r\|_2, \\
    d_{a}(\mathbf{h}_a, \mathbf{t}_a) &= \|\sin((\mathbf{h}_a + \mathbf{r}_a - \mathbf{t}_a)/2)\|_1
\end{align}
where $\| \cdot \|_1$, $ \| \cdot \|_2$ denote the $\ell_1$, $\ell_2$ norm. 
The final circular arc distance $d(\mathbf{h}, \mathbf{t})$ is the weighted sum of both radial and angular distances.
Note that, we can get $\mathbf{h}$ and $\mathbf{t}$ by concatenating both radial and angular vectors.
Such embedding model can then be pretrained individually through link prediction tasks, which helps us understand priors from KG into trainable embeddings. 
The regular negative sampling loss function $L$ is adopted for optimization:
\begin{gather}
L = -\log \sigma(\gamma - d(\mathbf{h}, \mathbf{t}))
- \sum_{i=1}^n \log \sigma(d(\mathbf{h}', \mathbf{t}') - \gamma)
\end{gather}
where $\gamma$ and $\sigma$ are marginal value and sigmoid function respectively, and $({h'}, r, {t'})$ is a negative triple.
Therefore, we can get diagnosis embedding matrix $\mathbf{E_{\mathcal{D}}}$ for condition codes. 

Note that we also experiment with a simpler objective function (i.e., bringing connected nodes closer while pushing unconnected nodes apart) as a replacement for the link prediction task. However, due to the complexity of the KG, the model fails to fully learn the positions of nodes (refer to the ablation study in section~\ref{sec:ablation}).

\subsection{Encode Graph-Enhanced Patient Embeddings}
The continual enhancement process unfolds in two steps: graph learning and proxy-task learning, which together refine the priors from the ``Knowledge Scholar'' by fully leveraging EHR data.
The proposed encoder-decoder framework not only emphasizes the comprehensive understanding of medical patterns but also seamlessly integrates local knowledge into the learning process, further capturing critical medical patterns for precise diagnosis.
The backbone of the encoder module contains graph learning and attention, which adjusts code-level embeddings and progressively refines them into comprehensive patient representations.

We construct a disease-complication graph $\mathcal{G}$ including both condition and laboratory codes. 
Compared to regular complication graphs involving only diseases, we have broader scope to explain complication in terms of shared abnormal lab tests in the same admission.
To represent the complications between two diseases via patient admissions, we add bidirectional edges $(c_i, c_j)$, $(c_i, c_m)$, and $(c_j, c_m)$ in graph $\mathcal{G}$, where $c_i, c_j \in \mathcal{D}$ are diagnostic codes and $c_m \in \mathcal{L}$ is a laboratory code.
% \yue{grammar and confusing} To describe the complication of two diseases via patient admissions, diagnostic codes $c_i, c_j \in \mathcal{D}$ and laboratory codes $c_m \in \mathcal{L}$, we add bidirectional edges $(c_i, c_j)$, $(c_i, c_m)$ and $(c_j, c_m)$ in graph $\mathcal{G}$.
Therefore, we then compute co-occurrence matrix 
$\mathbf{B} \in \mathbb{N}^{|\mathcal{C}| \times |\mathcal{C}|}$ and adjacent matrix 
$\mathbf{A} = (1-\varphi)\mathbf{B}+ \varphi \mathbf{I} \in \mathbb{R}^{|\mathcal{C}| \times |\mathcal{C}|}$ considering self loop based on EHR graph $\mathcal{G}$, where $\mathbf{I}$ is an identity matrix. 

A GNN with $L$ layers is adopted by EHR graph $\mathcal{G}$ and node features $\mathbf{E}$, where we initialize diseases by priors $\mathbf{E_{\mathcal{D}}}$ from KG.
Then, the hidden representation $\mathbf{H}^{(l)}$ can be calculated by input $\mathbf{H}^{(l-1)}$ through the $l$-th Convolution layer:
% \[
% \mathbf{H}^{(l+1)} = \text{ReLU}\left({\mathbf{A}}\mathbf{H}^{(l)}\mathbf{W}_{g}^{(l)}\right).
% \]
% Here, $\mathbf{H}^{(l)}$ and $\mathbf{H}^{(l+1)}$ are the input and output of the $l$th layer. $\mathbf{W}_{g}^{(l)}$ is the learnable weight of the $l$th layer.
After the last GNN layer, we obtain the hidden representation of diseases 
$\mathbf{X} = \textsc{GNN}(\mathbf{A}, \mathbf{E}) = \mathbf{H}^{(L)}$, which can be further refined through the bi-attention mechanism across codes and admissions for patient embedding $\mathbf{p}$:
\begin{align}
\mathbf{v}_\tau = \sum_{i=1}^{n} \alpha_{\tau}^i \mathbf{x}_i, \quad
\mathbf{p} = \sum_{\tau=1}^{T} \beta_{\tau} \tilde{\mathbf{v}}_{\tau},
\end{align}
where $\alpha_{\tau}^i$ and $\beta_{\tau}$ are attention scores computed as follows:
\begin{align}
\mathbf{z}_i = \tanh(\mathbf{W}_c \mathbf{x}_i), \quad 
\mathbf{r}_{\tau} = \tanh(\mathbf{W}_{v} \sigma(\mathbf{W}_{u} \mathbf{v}_{\tau}))
\end{align}
\begin{align}
\alpha_{\tau}^i = \frac{\exp(\mathbf{z}_i)}{\sum_{j=1}^{n} \exp(\mathbf{z}_j)}, \quad 
\beta_{\tau} = \frac{\exp(\mathbf{r}_{\tau})}{\sum_{\tau=1}^{T} \exp(\mathbf{r}_{\tau})}
\end{align}
$\mathbf{W}_{c}$, $\mathbf{W}_{u}$, and $\mathbf{W}_{v}$ are weighted matrices.
The attention score $\alpha_{\tau}$ measures the distribution of medical codes of admission, and $\beta$ measures the distribution across admissions.
Note that, we project each admission $\mathbf{v}_{\tau}$ to $\tilde{\mathbf{v}}_{\tau}$ for fitting the patient dimension through a single layer with $\mathbf{W}_{u}$. 

% The proposed encoder module ensures that both \yue{what does this mean? ``the specificity of individual medical encounters ''}the specificity of individual medical encounters and the comprehensive history of a patient’s health are reflected in the final patient embedding $\mathbf{p}$, laying a robust foundation for the subsequent self-supervised learning phase.

The proposed encoder module ensures that both the visit-level diagnostic information and the longitudinal history of a patient’s health are effectively captured in the final patient embedding $\mathbf{p}$, laying a robust foundation for the subsequent self-supervised learning phase.

\subsection{Stepwise Lab-Informed Pretraining} 

Most existing methods~\cite{ShangMXS19, LuRN23, poulain2024graph} adapt pretraining tasks from language models to exploit the whole EHR dataset, but fewer of them consider whether definitions are clinically meaningful. 
Motivated by this gap, we aim to harness laboratory insights to help predictive models better mimic the clinical reasoning process of a human doctor. 
Hence, we introduce a novel proxy-task learning paradigm that combine two clinically grounded steps, (1) \textbf{lab test assignment} and (2) \textbf{abnormality detection}, into the pretraining phase, thereby aligning the training process more closely with stepwise diagnostic reasoning.
To the best of our knowledge, this is the first attempt to explicitly incorporate lab test outcomes as labels during pretraining.

The EHR dataset for the proxy task learning is denoted as $\mathcal{S}'$ with $N'$ patients in total. 
Since those single-admission patients and final admission records of multi-admission patient lack labels regarding their next admission, we assume that the learned representations $\{\mathbf{p}_{1}, \mathbf{p}_{2}, \dots, \mathbf{p}_{N'}\}$ will be able to reflect their medical histories (i.e. historical lab test results).
Considering diverse patterns across different lab tests within the vocabulary $\mathcal{L} = \{l_1, l_2, \dots, l_{|\mathcal{L}|}\}$, we further categorize lab tests into several distinct clusters.
Note that, the categorical information of each test is documented in most EHR datasets. 
Here we take lab tests in MIMIC datasets as an example:
% \footnote{These categories have been already well documented and aligned in both MIMIC-III and MIMIC-IV.}:

\begin{enumerate}
    \item \textbf{Hematology} (\# items: 417)
    Analyze blood components like red and white blood cells to monitor conditions.
    \item \textbf{Chemistry} (\# items: 286)
    Evaluate chemical factors in blood and provide information about organ's function.
    \item \textbf{Blood Gas} (\# items: 35)
    Measure the levels of gas in blood to assess respiratory and metabolic functions.
\end{enumerate}

However, it still remains challenging to enable models to grasp the diagnostic reasoning process. 
 % \yue{I think the doctors first ask symptoms, not check history:}
Typically, a doctor first forms preliminary diagnostic hypotheses based on symptoms and medical history, orders relevant lab tests accordingly, and then confirms the diagnosis based on abnormal test results. 
This sequential reasoning underscores how to quantify conditional probabilities, which assign lab tests before observing abnormal outcomes, for our proxy-task learning module.

For better demonstration, we then separate the vocabulary $\mathcal{L}$ into $\mathcal{L}_1$, $\mathcal{L}_2$, and $\mathcal{L}_3$ in order and take single patient as an example.
Since we aim to decode embedding $\mathbf{p}$ into the probability distribution
$\mathbf{\hat{y}}: \mathbf{\hat{y}}_i = P(l_i \in \mathcal{L}_m|\mathbf{p})$ for each category $m$, it is intuitive that the equation can be parameterized by a multi-layer perceptron (MLP) for estimation $\mathbf{\hat{y}}=\sigma(\text{MLP}(\mathbf{p}))$.
However, such na\"{i}ve method has limitation in mimicking the conditional probability for both assignment or abnormality step.
Assuming assigned and abnormal lab results are denoted as $\mathbf{\hat{y}}^{a,m}$ and $\mathbf{\hat{y}}^{b,m}$ respectively for certain category $m$, the abnormal lab results $\mathbf{\hat{y}}^{b,m}$ is equivalent to a joint probability $\text{P}\left(p^{a}_{l_i} \in \mathcal{L}_{m},p^{b}_{l_i} \in \mathcal{L}_{m}|\mathbf{p}\right)$ regarding assignment as priors, which can be represented as follow with given $\mathbf{p}$:
\begin{align}\nonumber
    \mathbf{\hat{y}}^{b,m}
    &= \text{P}\left(
    p^{a}_{l_i} \in \mathcal{L}_{m},
    p^{b}_{l_i} \in \mathcal{L}_{m}|\mathbf{p}
    \right)\\
    &= \text{P}\left(
    p^{b}_{l_i} \in \mathcal{L}_{m}|
    p^{a}_{l_i} \in \mathcal{L}_{m},\mathbf{p}
    \right) \times \text{P}\left(
    p^{a}_{l_i} \in \mathcal{L}_{m}|\mathbf{p}
    \right)
\end{align}
where $i$ represent the index within the vocabulary $\mathcal{L}$, so that $p^{a}_{l_i}$ and $p^{b}_{l_i}$ are probabilities of the $i$-th assigned and abnormal lab results.
Even though we separate our target $\mathbf{\hat{y}}^b$ by two consecutive conditional components, it is still difficult to parameterize 
$\text{P}\left(
p^{b}_{l_i} \in \mathcal{L}_{m}|
p^{a}_{l_i} \in \mathcal{L}_{m},\mathbf{p}
\right)$
for estimation.
Hence, we adopt the constraint relaxation method and parameterize
$\text{P}\left(
p^{b}_{l_i} \in \mathcal{L}_{m}|\mathbf{p}
\right)$
as the lower bound approximation:
\begin{align}
    \text{P}\left(
    p^{b}_{l_i} \in \mathcal{L}_{m}|
    p^{a}_{l_i} \in \mathcal{L}_{m},\mathbf{p}
    \right) &\geq \text{P}\left(
    p^{b}_{l_i} \in \mathcal{L}_{m}|\mathbf{p}
    \right) 
    \end{align}
    \begin{align}
    \text{P}\left(
    p^{a}_{l_i} \in \mathcal{L}_{m}|\mathbf{p}
    \right) &= \sigma(\mathbf{w}_a\mathbf{p})_{l_i}\\
    % \quad
    \text{P}\left(
    p^{b}_{l_i} \in \mathcal{L}_{m}|\mathbf{p}
    \right) &= \sigma(\mathbf{w}_b\mathbf{p})_{l_i}\\
    \mathbf{\hat{y}}^{b,m}_i 
    &= [\sigma(\mathbf{w}^m_a\mathbf{p}) \otimes
    \sigma(\mathbf{w}^m_b\mathbf{p})]_{l_i}
\end{align}
\begin{align}
    L^m = \; &\omega \times \text{BCE}(\mathbf{\hat{y}}^{a,m}, \mathbf{y}^{a,m}) \\\notag
    & + (1-\omega) \times \text{BCE}(\mathbf{\hat{y}}^{b,m}, \mathbf{y}^{b,m})
\end{align}
where $\mathbf{w}^m_a, \mathbf{w}^m_b \in \mathbb{R}^{|\mathcal{L}_m| \times p}$ decodes patient embedding $\mathbf{p}$ to the probability of the laboratory codes for assignment and abnormality respectively.
In $L^m$, $\text{BCE}(\cdot)$ denotes the binary cross-entropy loss and $\omega$ is the hyperparameter scaling two losses. 
$\mathbf{y}^{a,m}$ and $\mathbf{y}^{b,m}$ are ground truth for prediction.
Note that, as a self-supervised learning problem, such design help model learn the conditional relationship between both steps, which further augment the robustness of learned parameters.
Moreover, to ensure the lab embeddings are well-aligned with each category, the proxy task will be trained in two phrases:
% Hence, we separate lab tests into $\mathcal{L}_1$, $\mathcal{L}_2$, and $\mathcal{L}_3$ in order, and predict lab results $\hat{y}_{i,j} = \text{Decoder}(\mathbf{p})$ based on both single-admission and multi-admission data in training set, where $y_{i,j}=1$ means lab code $l_j$ shows abnormal results, and a multi-layer perceptron (MLP) is used as decoder for learning patterns in different lab categories.
% We only focus on abnormal lab tests across admissions, so the ground truth $[l_1, l_2,...,l_{|\mathcal{L}|}]$ are labeled as multi-hot vectors.
% To ensure the lab embeddings are well-aligned with each category, the proxy task is designed to be completed in two steps:

\begin{itemize}
    \item \textbf{Joint Training.} We get preliminary trained parameters in encoder $\Theta_{e}$ and optimize prediction through the average loss across three categories, where we can adjust patient embedding $\tilde{\mathbf{p}}$ through updated parameters.
    % \begin{align}
    % L_{j} = \frac{1}{|\mathcal{Y}|} \sum_{i=1}^{|\mathcal{Y}|} \text{BCE}(\mathbf{\hat{y}}_{i}, \mathbf{y}_{i}), \:  
    % \mathcal{Y}=\{\mathcal{L}_1,\mathcal{L}_2,\mathcal{L}_3\} 
    % \end{align}
    % where $\mathbf{y}_{i}$ and $\mathbf{\hat{y}}_{i}$ are ground truth and prediction of lab results and $i$ means the $i$-th lab category.
    \item \textbf{Individual Training.} We individually update parameters in decoder $\Theta_{d}$ after maintaining $\Theta_{e}$ unchanged given from the former phase, where we get refined lab embeddings $\mathbf{\tilde{l}_1}$, $\mathbf{\tilde{l}_2}$, and $\mathbf{\tilde{l}_3}$ through updated decoder.
    % \begin{align}
    % L_{i} = \text{BCE}(\mathbf{\hat{y}}_{i}, \mathbf{y}_{i}), \: i=1,2,3
    % \end{align}
\end{itemize}

After two training phases, we get pretrained parameter $\Theta = [\Theta_{d};\Theta_{e}]$ for the predictive model.
Similar to the real scenario, doctors always provide diagnoses relying on their historical conditions and abnormal lab results.
By training model to understand such underlying laboratory pattern, it provides a pathway to deeply integrate additional laboratory knowledge into the model, ensuring that the predictions are not only accurate but also clinically meaningful.

\subsection{Fine-Tuning and Inference}
The proposed model can be trained with EHR data either with or without lab tests, making it adaptable to data missing desired features. If training with only diagnostic information, we can directly compute the estimated output $\hat{\mathbf{y}}'$ and the downstream loss $\mathcal{L}'$ for optimization:
\begin{align}
\hat{\mathbf{y}}'_{\text{direct}} &= \sigma(\mathbf{W} \mathbf{p}) \in \mathbb{R}^o \\
L'_{\text{direct}} &= \text{Loss}(\mathbf{y}', \hat{\mathbf{y}}')
\end{align}

If lab features are included, the model leverages pretrained parameters $\Theta$ to get both adjusted patient embedding $\tilde{\mathbf{p}}$ and lab embeddings $\mathbf{\tilde{l}_1}$, $\mathbf{\tilde{l}_2}$, and $\mathbf{\tilde{l}_3}$. The downstream outputs $\hat{\mathbf{y}}'$ and fine-tuning loss $\mathcal{L}'$ are then calculated as follows:
\begin{align}
\hat{\mathbf{y}}'_{\text{finetune}} &= \sigma(\mathbf{W} [\mathbf{\tilde{l}_1}: \mathbf{\tilde{l}_2}: \mathbf{\tilde{l}_3}] | 
\tilde{\mathbf{p}}) \in \mathbb{R}^o \\
L'_{\text{finetune}} &= \text{Loss}(\mathbf{y}', \hat{\mathbf{y}}' | \Theta)
\end{align}
Here, $\mathbf{W} \in \mathbb{R}^{o \times p}$ is the weight matrix and $[:]$ means concatenation. 
Note that $\mathbf{y}'$, $\sigma$, $o$, and $\mathcal{L}'$ depend on the specific tasks. During optimization with backpropagation, the embedding matrix and parameters remain learnable.

\section{Experiments}

\subsection{Experiment Setting}

\subsubsection{Datasets}

To evaluate our proposed model, we utilize two public and widely-used EHR datasets: MIMIC-III~\cite{johnson2016mimic} and MIMIC-IV~\cite{johnson2020mimic}.
MIMIC-III includes 7,493 patients with multiple visits ($T \geq 2$) recorded between 2001 and 2012, while MIMIC-IV contains 85,155 such patients from 2008 to 2019.
To ensure a comparable sample size and avoid temporal overlap with MIMIC-III, we randomly sample 10,000 patients from MIMIC-IV who were admitted between 2013 and 2019.
Table~\ref{tab:stats_mimic} summarizes the basic statistics of the MIMIC-III and MIMIC-IV datasets.
We divide both datasets into training, validation, and test sets by patient: 6,000/493/1,000 patients for MIMIC-III and 8,000/1,000/1,000 for MIMIC-IV, respectively.
For each patient, the most recent visit is used as the prediction target, with all previous visits serving as input features.
All diagnoses are encoded using the ICD-9-CM coding system.

\begin{table}[t]
    \centering
    \small
    \caption{Statistics of MIMIC-III and MIMIC-IV datasets}
    \label{tab:stats_mimic}
    \begin{tabular}{lcc}
    \toprule
    \textbf{Dataset} & \textbf{MIMIC-III} & \textbf{MIMIC-IV} \\ \midrule
    \# patients & 7,493 & 10,000 \\
    Max. \# visit & 42 & 55 \\
    Avg. \# visit & 2.66 & 3.66 \\ \midrule
    \# codes & 4,880 & 6,102 \\
    Max. \# codes per visit  & 39 & 50 \\
    Avg. \# codes per visit  & 13.06 & 13.38 \\ 
    \bottomrule
    \end{tabular}
\end{table}

\subsubsection{Baselines}
To evaluate the performance of our proposed model, we selected the following 10 EHR models as comparison methods:
(1) \textbf{RNN/CNN-based models}: RETAIN~\cite{ChoiBSKSS16} and Timeline~\cite{BaiZEV18}.
(2) \textbf{Graph/KG-based models}: GCT~\cite{ChoiXLDFXD20}, CGL~\cite{LuRCKN21}, SeqCare~\cite{xu2023seqcare}, and RAM-EHR~\cite{xu2024ram}. 
(3) \textbf{Transformer-based models}: G-BERT~\cite{ShangMXS19}, HiTANet~\cite{LuoYXM20}, Med-BERT~\cite{rasmy2021med}, and GT-BEHRT~\cite{poulain2024graph}.

% drop baselines: Dipole~\cite{MaCZYSG17}, Deepr~\cite{NguyenTWV17}, KAME, GRAM
Note that although GraphCare~\cite{jiang2023graphcare} is another KG-based method, it is not included as a baseline in experiments because its defined KGs cannot be directly transferred from CCS codes to ICD-9 codes.
However, we feed our diagnosis KG into GraphCare framework for comparison as shown in Table~\ref{tab:ablation}.

\subsubsection{Tasks \& Evaluation Metrics.}
We conduct our experiments on two tasks as downstream examples for clinical prediction, following the similar settings as in previous studies~\cite{ChoiBSSS17, LuRCKN21}:

\begin{enumerate}
    \item \textbf{Diagnosis Prediction.} This task involves multi-label prediction, where the goal is to predict all condition codes for the next admission based on admission history. 
    \item \textbf{Heart Failure Prediction (HF).} This is a binary classification task that predicts whether patients will be diagnosed with heart failure in their next admission.
\end{enumerate}

Given the label imbalance in EHR data, we use the weighted $F_1$ score (w-$F_1$) and recall at k ($R@k$) as evaluation metrics for diagnosis prediction and use the area under the ROC curve (AUC) and the F\textsubscript{1} score for HF prediction.

\subsubsection{Implementation Details}
During the transformation of textual information into triples via LLMs, each inference undergoes iterative prompting and re-read prompting strategies with two and one rounds, respectively.
Polar-space hierarchical KG embeddings are used as node features in the graph learning module, with vector size $2k = 2000$ ($k = 1000$ for both hierarchical and semantic components). The disease-lab graph is modeled using two GAT convolution layers with hidden dimensions of 256, and default settings elsewhere. Attention modules adopt dimensions $a = 256$ and $b = 256$, with a dropout rate of 0.2 applied to each.
Lab test embeddings are generated using three decoders, each implemented as a three-layer MLP with hidden sizes of 256 and 128, and a uniform dropout rate of 0.4 across all layers for both pretraining and direct training. A two-layer classifier with 256 hidden units and a dropout rate of 0.5 is used to transform concatenated features into logits for downstream tasks.
KG embeddings are trained for 180{,}000 steps to capture hierarchical and semantic relations. Proxy-task learning is performed with 10 epochs for both joint and individual training. Diagnosis and heart failure predictions use 500 and 50 training epochs, respectively. The Adam optimizer with a decaying learning rate is employed throughout, with an initial learning rate of 0.001 for both pretraining and finetuning. All experiments are conducted using Python 3.10 and PyTorch 2.3.1 with CUDA 12.4 on a system equipped with two AMD EPYC 9254 24-Core processors, 528 GB RAM, and four NVIDIA L40S GPUs.

\subsection{Prediction Results}

\begin{table*}[t]
\centering
\small
\caption{Prediction Results on MIMIC-III and MIMIC-IV for Diagnosis (Task 1) and HF (Task 2) Prediction. We report the average performance (\%) and standard deviation (in bracket) of each model over 10 runs. The best results are highlighted.}
\label{tab:main_result}
\begin{tabular}{l ccc ccc cc cc}
\toprule
\multirow{3}{*}{\textbf{Model}} 
& \multicolumn{6}{c}{\textbf{Task 1: Diagnosis Prediction}} 
& \multicolumn{4}{c}{\textbf{Task 2: Heart Failure Prediction}} \\
% \cmidrule(lr){2-7} \cmidrule(lr){8-11}
& \multicolumn{3}{c}{\textbf{MIMIC-III}} 
& \multicolumn{3}{c}{\textbf{MIMIC-IV}}
& \multicolumn{2}{c}{\textbf{MIMIC-III}}
& \multicolumn{2}{c}{\textbf{MIMIC-IV}} \\
\cmidrule(lr){2-4} \cmidrule(lr){5-7} \cmidrule(lr){8-9} \cmidrule(lr){10-11}
& \textbf{w-F1} & \textbf{R@10} & \textbf{R@20}
& \textbf{w-F1} & \textbf{R@10} & \textbf{R@20}
& \textbf{AUC} & \textbf{F1}
& \textbf{AUC} & \textbf{F1} \\
\midrule
RETAIN
& 18.37$_{(0.8)}$ & 32.12$_{(0.8)}$ & 32.54$_{(0.6)}$
& 23.11$_{(0.8)}$ & 37.32$_{(0.8)}$ & 40.15$_{(0.6)}$
& 83.21$_{(0.3)}$ & 71.32$_{(0.2)}$
& 84.14$_{(0.3)}$ & 71.23$_{(0.2)}$ \\
Timeline
& 20.46$_{(0.2)}$ & 30.73$_{(0.1)}$ & 34.83$_{(0.1)}$
& 23.76$_{(0.2)}$ & 37.89$_{(0.1)}$ & 40.87$_{(0.1)}$
& 82.34$_{(0.3)}$ & 71.03$_{(0.2)}$
& 83.45$_{(0.3)}$ & 72.30$_{(0.2)}$ \\
GCT
& 20.66$_{(0.2)}$ & 32.73$_{(0.2)}$ & 35.44$_{(0.2)}$
& 24.16$_{(0.2)}$ & 38.24$_{(0.2)}$ & 41.65$_{(0.2)}$
& 82.08$_{(0.3)}$ & 70.35$_{(0.2)}$
& 84.80$_{(0.3)}$ & 69.52$_{(0.2)}$ \\
CGL
& 22.63$_{(0.2)}$ & 33.64$_{(0.3)}$ & 37.87$_{(0.2)}$
& 25.74$_{(0.2)}$ & 39.23$_{(0.3)}$ & 42.67$_{(0.2)}$
& 84.19$_{(0.2)}$ & 71.77$_{(0.1)}$
& 87.91$_{(0.2)}$ & 70.71$_{(0.3)}$ \\
SeqCare
& 21.78$_{(0.1)}$ & 34.17$_{(0.2)}$ & 36.46$_{(0.3)}$
& 24.39$_{(0.1)}$ & 38.42$_{(0.2)}$ & 41.62$_{(0.3)}$
& 81.55$_{(0.2)}$ & 68.78$_{(0.1)}$
& 85.55$_{(0.2)}$ & 69.82$_{(0.4)}$ \\
RAM-EHR
& 23.71$_{(0.1)}$ & 37.97$_{(0.2)}$ & 40.18$_{(0.2)}$
& 27.01$_{(0.1)}$ & 42.86$_{(0.2)}$ & 46.92$_{(0.2)}$
& 82.88$_{(0.1)}$ & 72.03$_{(0.1)}$
& 84.80$_{(0.2)}$ & 72.34$_{(0.5)}$ \\
G-BERT
& 22.28$_{(0.3)}$ & 35.62$_{(0.2)}$ & 36.46$_{(0.2)}$
& 25.12$_{(0.3)}$ & 39.91$_{(0.2)}$ & 43.25$_{(0.2)}$
& 81.50$_{(0.2)}$ & 71.18$_{(0.1)}$
& 85.76$_{(0.2)}$ & 72.88$_{(0.1)}$ \\
HiTANet
& 23.15$_{(0.2)}$ & 34.68$_{(0.3)}$ & 35.97$_{(0.1)}$
& 24.53$_{(0.2)}$ & 38.42$_{(0.3)}$ & 41.89$_{(0.1)}$
& 85.13$_{(0.1)}$ & 73.15$_{(0.2)}$
& 86.34$_{(0.4)}$ & 71.35$_{(0.2)}$ \\
Med-BERT
& 21.68$_{(0.1)}$ & 33.47$_{(0.1)}$ & 36.53$_{(0.1)}$
& 23.58$_{(0.1)}$ & 36.79$_{(0.1)}$ & 40.45$_{(0.1)}$
& 81.36$_{(0.1)}$ & 69.54$_{(0.1)}$
& 83.61$_{(0.5)}$ & 70.46$_{(0.1)}$ \\
GT-BEHRT
& 23.92$_{(0.1)}$ & 38.64$_{(0.2)}$ & 39.97$_{(0.1)}$
& 26.97$_{(0.3)}$ & 43.07$_{(0.4)}$ & 47.19$_{(0.2)}$
& 83.24$_{(0.1)}$ & 74.12$_{(0.1)}$
& 87.43$_{(0.3)}$ & 72.26$_{(0.2)}$ \\
\midrule
\textbf{\system{}}
& \textbf{25.37}$_{(0.3)}$ & \textbf{40.52}$_{(0.2)}$ & \textbf{41.86}$_{(0.3)}$
& \textbf{29.87}$_{(0.3)}$ & \textbf{45.66}$_{(0.2)}$ & \textbf{51.73}$_{(0.3)}$
& \textbf{86.53}$_{(0.1)}$ & \textbf{75.35}$_{(0.1)}$
& \textbf{90.32}$_{(0.2)}$ & \textbf{73.54}$_{(0.1)}$ \\
\bottomrule
\end{tabular}
\end{table*}

\begin{table}[t]
    \centering
    \caption{Supplementary Prediction Results on MIMIC-III for Real-time and Non-chronic Diagnosis Prediction.}
    \label{tab:mimic3_supplement}
    \begin{tabular}{@{}lccccc@{}}
        \toprule
         & \multicolumn{3}{c}{\textbf{RT Diagnosis}} & \multicolumn{2}{c}{\textbf{NC Diagnosis}} \\
        % \cmidrule(lr){2-4} \cmidrule(lr){5-6}
        \textbf{Models} & \textbf{w-}$\mathbf{F_1}$ & \textbf{R@10} & \textbf{R@20} & \textbf{w-}$\mathbf{F_1}$ & \textbf{R@10} \\
        \midrule
        RETAIN   & 19.56$_{(0.2)}$ & 32.18$_{(0.3)}$ & 33.67$_{(0.3)}$ & 18.22$_{(0.5)}$ & 30.15$_{(0.4)}$ \\
        Timeline & 21.12$_{(0.2)}$ & 33.05$_{(0.3)}$ & 34.23$_{(0.3)}$ & 18.04$_{(0.3)}$ & 29.58$_{(0.4)}$ \\
        GCT      & 22.54$_{(0.2)}$ & 35.67$_{(0.3)}$ & 37.10$_{(0.4)}$ & 18.95$_{(0.3)}$ & 30.08$_{(0.3)}$ \\
        CGL      & 24.13$_{(0.1)}$ & 38.21$_{(0.3)}$ & 39.64$_{(0.4)}$ & 19.45$_{(0.3)}$ & 30.52$_{(0.5)}$ \\
        SeqCare  & 23.21$_{(0.1)}$ & 36.51$_{(0.1)}$ & 37.75$_{(0.3)}$ & 19.13$_{(0.2)}$ & 30.34$_{(0.4)}$ \\
        RAM-EHR  & 24.19$_{(0.1)}$ & 38.74$_{(0.2)}$ & 41.81$_{(0.2)}$ & 19.56$_{(0.3)}$ & 31.27$_{(0.4)}$ \\
        G-BERT   & 24.35$_{(0.3)}$ & 37.12$_{(0.5)}$ & 39.35$_{(0.4)}$ & 19.26$_{(0.2)}$ & 31.96$_{(0.3)}$ \\
        HiTANet  & 25.67$_{(0.3)}$ & 39.72$_{(0.4)}$ & 41.85$_{(0.4)}$ & 18.81$_{(0.3)}$ & 31.08$_{(0.4)}$ \\
        Med-BERT & 22.74$_{(0.3)}$ & 35.27$_{(0.4)}$ & 36.54$_{(0.4)}$ & 17.94$_{(0.3)}$ & 30.84$_{(0.4)}$ \\
        GT-BEHRT & 24.53$_{(0.4)}$ & 40.31$_{(0.2)}$ & 41.45$_{(0.3)}$ & 19.93$_{(0.3)}$ & 33.53$_{(0.4)}$ \\
        \midrule
        \textbf{\system{}}\;  & \textbf{26.89}$_{(\textbf{0.3})}$ & \textbf{42.86}$_{(\textbf{0.1})}$ & \textbf{43.15}$_{(\textbf{0.2})}$ & \textbf{21.69}$_{(\textbf{0.2})}$ & \textbf{34.35}$_{(\textbf{0.4})}$ \\
        \bottomrule
    \end{tabular}
\end{table}

Table~\ref{tab:main_result} demonstrates that \system{} consistently outperforms existing baselines across both diagnosis and heart failure prediction tasks on both MIMIC-III and MIMIC-IV datasets, which underlines the efficacy of external priors from clinical KGs for prediction.
For diagnosis prediction, we can observe that \system{} boosts the weighted F\textsubscript{1} score by approximately 6\% relative to the best baseline on MIMIC-III. 
Similar trends can be observed in the R@10 and R@20 metrics, where \system{} surpasses the strongest baselines by around 4--5\%.
The improvements on MIMIC-IV are even more pronounced, since \system{} reaches a w-$F_1$ of 29.87\%, an improvement of nearly 11\%. 
The R@10 and R@20 metrics follow suit, with \system{} outperforming the best baseline by about 6\% and 10\%, respectively. 
Regarding the HF prediction task, the absolute gains are somewhat smaller but remain consistent. 

Table~\ref{tab:mimic3_supplement} shows the results of two modified tasks, real-time and non-chronic diagnoses, on the MIMIC-III dataset:
\begin{enumerate}
    \item \textbf{Real-time (RT) Diagnosis}: To fairly compare performance of all baselines, we feed lab tests information for all baselines and proposed model, which not only augments representation for diagnosis prediction but also mimics real-time diagnosis of physician within the beginning of an admission. We can observe that the proposed model still outperforms all baselines, which has even a larger improvement on w-F\textsubscript{1} over regular diagnosis. We suppose that \system{}, which originally relies on lab tests for auxiliary diagnosis, will be more suitable for lab features input than other baselines.
    \item \textbf{Non-chronic (NC) Diagnosis}: Since MIMIC-III mostly collects ICU data, we remove all chronic diseases and focus on predicting acute diseases for future visits. 
    Specifically, there are 142 codes removed by the chronic condition indicator~\footnote{https://hcup-us.ahrq.gov/toolssoftware/chronic/chronic.jsp} for ICD-9-CM, and most of them frequently appears in admission records. 
    Therefore, all models suffer performance degradations compared to the regular setting.
    We can observe that \system{} still achieves superior performance, but all baselines and proposed model get relatively low values of metrics compared to regular diagnosis. 
    We conjecture that the removal of chronic diseases in admissions skews the distribution, making rare diseases more prominent and, consequently, making the task more challenging.
\end{enumerate}

Overall, these results suggest that the dual integration of public knowledge from clinical knowledge graphs and individualized patient data is crucial. 
By leveraging both external priors and patient-specific nuances, \system{} not only bridges the gap between structured medical information and dynamic EHR data but also consistently outperforms models that rely on either temporal learning, medical ontology, or Transformer-based architectures alone.

\subsection{Ablation Study}
\label{sec:ablation}

\begin{table}[t]
    \centering
    % \small
    \caption{Ablation Results of the Variants of \system{}.}
    \label{tab:ablation}
    \begin{tabular}{lccccc}
    \toprule
     & \multicolumn{3}{c}{\textbf{Diagnosis Prediction}} & \multicolumn{2}{c}{\textbf{HF Prediction}} \\
    \textbf{Model} & \textbf{w-F1} & \textbf{R@10} & \textbf{R@20} & \textbf{AUC} & \textbf{F1} \\
    \midrule
    \system{}$_{a}$ & 21.21$_{(0.2)}$ & 35.25$_{(0.3)}$ & 36.46$_{(0.3)}$ & 80.75$_{(0.4)}$ & 69.81$_{(0.3)}$ \\
    \system{}$_{b}$ & 22.32$_{(0.3)}$ & 36.28$_{(0.3)}$ & 36.95$_{(0.4)}$ & 81.66$_{(0.3)}$ & 70.01$_{(0.3)}$ \\
    \system{}$_{c}$ & 23.15$_{(0.3)}$ & 37.78$_{(0.4)}$ & 38.79$_{(0.4)}$ & 83.16$_{(0.3)}$ & 72.73$_{(0.3)}$ \\
    \system{}$_{d}$ & 22.49$_{(0.3)}$ & 34.86$_{(0.3)}$ & 35.79$_{(0.4)}$ & 80.90$_{(0.3)}$ & 69.24$_{(0.3)}$ \\
    \system{}$_{e}$ & 24.58$_{(0.3)}$ & 39.78$_{(0.4)}$ & 40.46$_{(0.4)}$ & 83.92$_{(0.3)}$ & 73.62$_{(0.3)}$ \\
    \system{}$_{f}$ & 24.96$_{(0.3)}$ & 40.16$_{(0.3)}$ & 40.87$_{(0.4)}$ & 85.81$_{(0.3)}$ & 73.96$_{(0.3)}$ \\
    PrimeKG$_{g}$ & 23.74$_{(0.3)}$ & 38.19$_{(0.3)}$ & 39.52$_{(0.4)}$ & 84.73$_{(0.3)}$ & 72.85$_{(0.3)}$ \\
    GraphCare$_{h}$ & 24.14$_{(-)}$ & 39.89$_{(-)}$ & 40.14$_{(-)}$ & 83.73$_{(-)}$ & \textbf{76.28}$_{(-)}$ \\
    \midrule
    \textbf{\system{}} & \textbf{25.37}$_{(\textbf{0.3})}$ & \textbf{40.52}$_{(\textbf{0.3})}$ & \textbf{41.86}$_{(\textbf{0.3})}$ & \textbf{86.53}$_{(\textbf{0.3})}$ & 75.35$_{(\textbf{0.3})}$ \\
    \textit{- w/o lab} & 24.21$_{(0.3)}$ & 39.47$_{(0.3)}$ & 40.35$_{(0.3)}$ & 85.73$_{(0.3)}$ & 73.92$_{(0.3)}$ \\
    \bottomrule
    \end{tabular}
\end{table}

To evaluate the effectiveness of involved modules in \system{}, we compare performance of variants with some modules removed or replaced. We use both future diagnosis and HF prediction tasks on MIMIC-III as examples:

\begin{enumerate}
    \item \textbf{\system{}\textsubscript{a}}: Replacing the embedding of the Diagnosis KG with \texttt{GloVe}, as used in GRAM~\cite{ChoiBSSS17}, to only focus on hierarchical information for diseases.
    \item \textbf{\system{}\textsubscript{b}}: Replacing the embedding of the Diagnosis KG with \texttt{text-embedding-3-large}~\cite{embedding2024} to only retrieve semantic information for diseases. 
    \item \textbf{\system{}\textsubscript{c}}: Replacing the link prediction tasks with minimizing distances among connected codes and maximizing distances among unconnected ones. 
    \item \textbf{\system{}\textsubscript{d}}: Removing the Graph Learning module and directly feeding KG embeddings into the Attention layer.
    \item \textbf{\system{}\textsubscript{e}}: Replacing EHR graph by a simplified graph with disease complications, excluding lab test nodes.
    \item \textbf{\system{}\textsubscript{f}}: Replacing Attention by GRU modules.
    \item \textbf{PrimeKG\textsubscript{g}}: 
    Using PrimeKG as the KG for initialization of disease embeddings by mapping identifier from MONDO to ICD-9-CM.
    \item \textbf{GraphCare\textsubscript{h}}: Feeding learned embeddings from the Diagnosis KG into the predictor used by Graphcare aligning to the same encoding system. 
\end{enumerate}

% Even though the F1 score of GraphCare\textsubscript{h} surpass \system{}, the much higher computational cost (i.e. running time and GPU coverage) cannot be ignored, which shows that the effectiveness of considering multi-aspect information other than fine-grained graphs for accurate diagnosis.

The results of ablation variants are shown in Table~\ref{tab:ablation}.
\system{} consistently outperforms its variants, validating the effectiveness of each module.
The comparison with \system{}\textsubscript{a-b} shows that both hierarchical and semantic information in the KG contribute to performance, even though each alone provides some benefit.
Results from \system{}\textsubscript{d-f} demonstrate the importance of graph learning and attention for representation refinement, with lab tests further enhancing performance.
The substantial gain over PrimeKG\textsubscript{g} highlights the value of a task-specific KG with relevant entities.
Although GraphCare\textsubscript{h} achieves higher F1, its higher computational cost underscores the efficiency of leveraging multi-aspect information beyond fine-grained graphs.
The model \textit{w/o lab} using modified GNN and pretraining tasks in Sherbet~\cite{LuRN23}, still achieves superiority over baselines, demonstrating the generalizability of \system{}.

\begin{figure}[t!]
    \centering
    \includegraphics[width=0.48\textwidth]{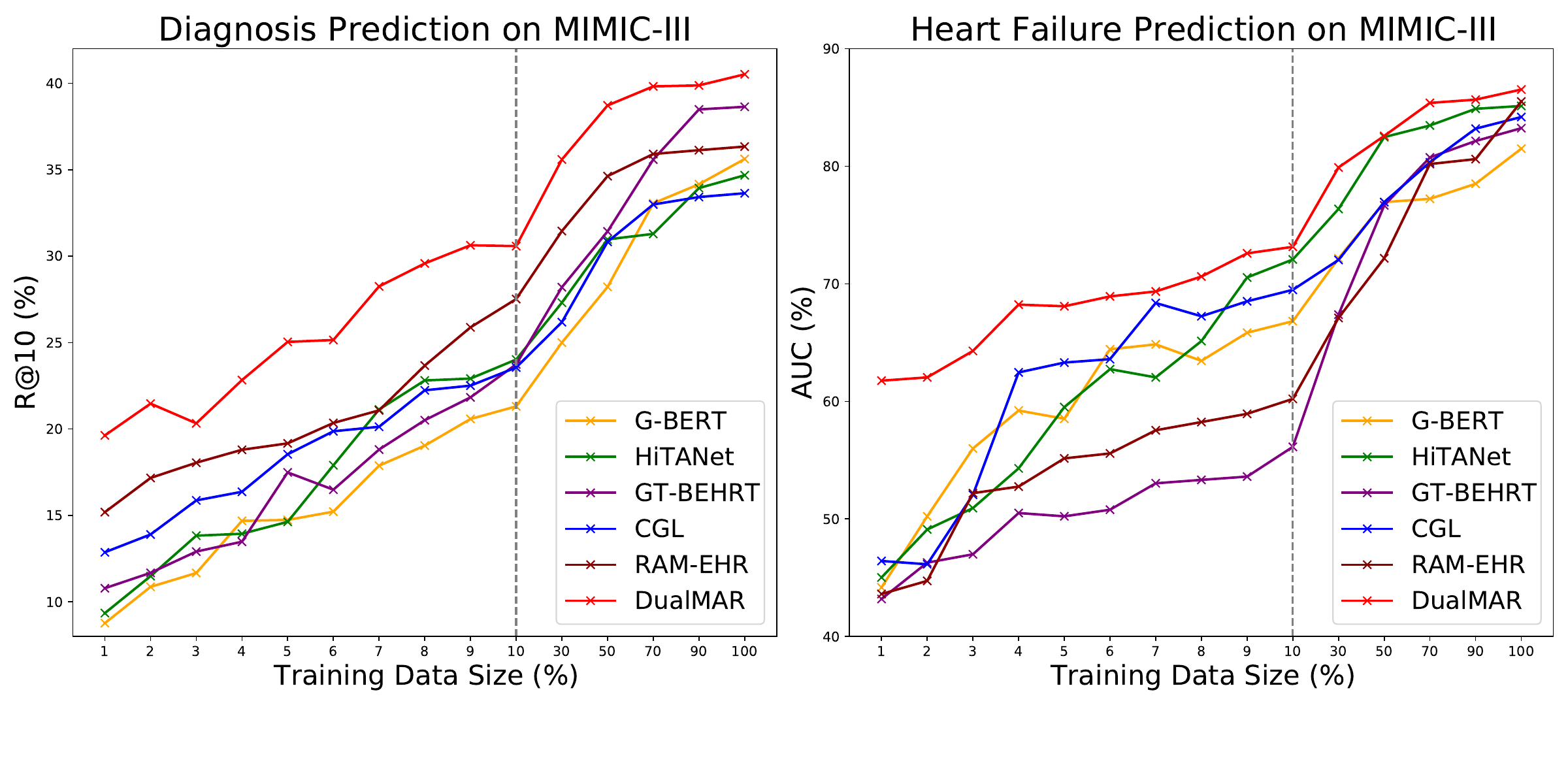}
    \caption{Performance by EHR Training Data Sizes. Values on the x-axis indicate \% of the entire training data. The dotted lines separate two ranges: [1, 10] and [10, 100]~(\%).}
    \label{fig:ehr_size}
\end{figure}

\subsection{Case Study}

\subsubsection{Effect of EHR Data Size}
By incorporating the diagnosis KG as medical prior knowledge before the training phase, \system{} is expected to exhibit greater robustness to variations in training data size. 
Hence, it is crucial to evaluate how well models can provide accurate diagnoses with limited labeled data, both with and without prior knowledge. 
Figure~\ref{fig:ehr_size} presents the results for R@10 in Diagnosis Prediction and AUC in heart failure prediction. As shown, all models experience performance degradation as the amount of labeled data decreases, especially when less than 10\% of the training set is available. 
However, \system{} consistently outperforms other baselines, demonstrating that knowledge injection from the KG effectively enhances information utilization, even in scenarios with scarce labeled data.

\subsubsection{Diagnosis for Low-Prevalence Diseases}
We also design experiments focusing on predicting less frequent condition codes in MIMIC-III, similar to low-prevalence medical coding tasks. 
% We conduct two subtasks:
% \begin{itemize}
%     \item \texttt{Rare-20}: Predicting 200 ICD codes that occur fewer than 20 times in MIMIC-III.
%     \item \texttt{Rare-10}: Predicting 100 ICD codes that occur fewer than 10 times in MIMIC-III.
% \end{itemize}
We select and predict 100 ICD codes that occur fewer than 10 times in the dataset. 
We follow the previous settings which excludes records without low-prevalence diseases to stabilize training. 
We use recall at 5/8/15/20 for evaluation.
Figure~\ref{fig:rare_diagnosis} compares \system{} with baselines using ontology or transformer architectures. 
\system{} outperforms other selected methods on R@5, R@8, R@15, and R@20.
Notably, CGL and G-BERT show significant improvements, indicating that external knowledge bases can aid in diagnosing less-frequent conditions. 
This underscores the importance of incorporating medical ontologies or complex KGs to enhance model resilience against EHR data limitations.

\subsubsection{Advanced Analysis of KG Embeddings} 
\begin{figure}[t!]
    \centering
    \begin{subfigure}{0.48\linewidth}
        \centering
        \includegraphics[width=\linewidth]{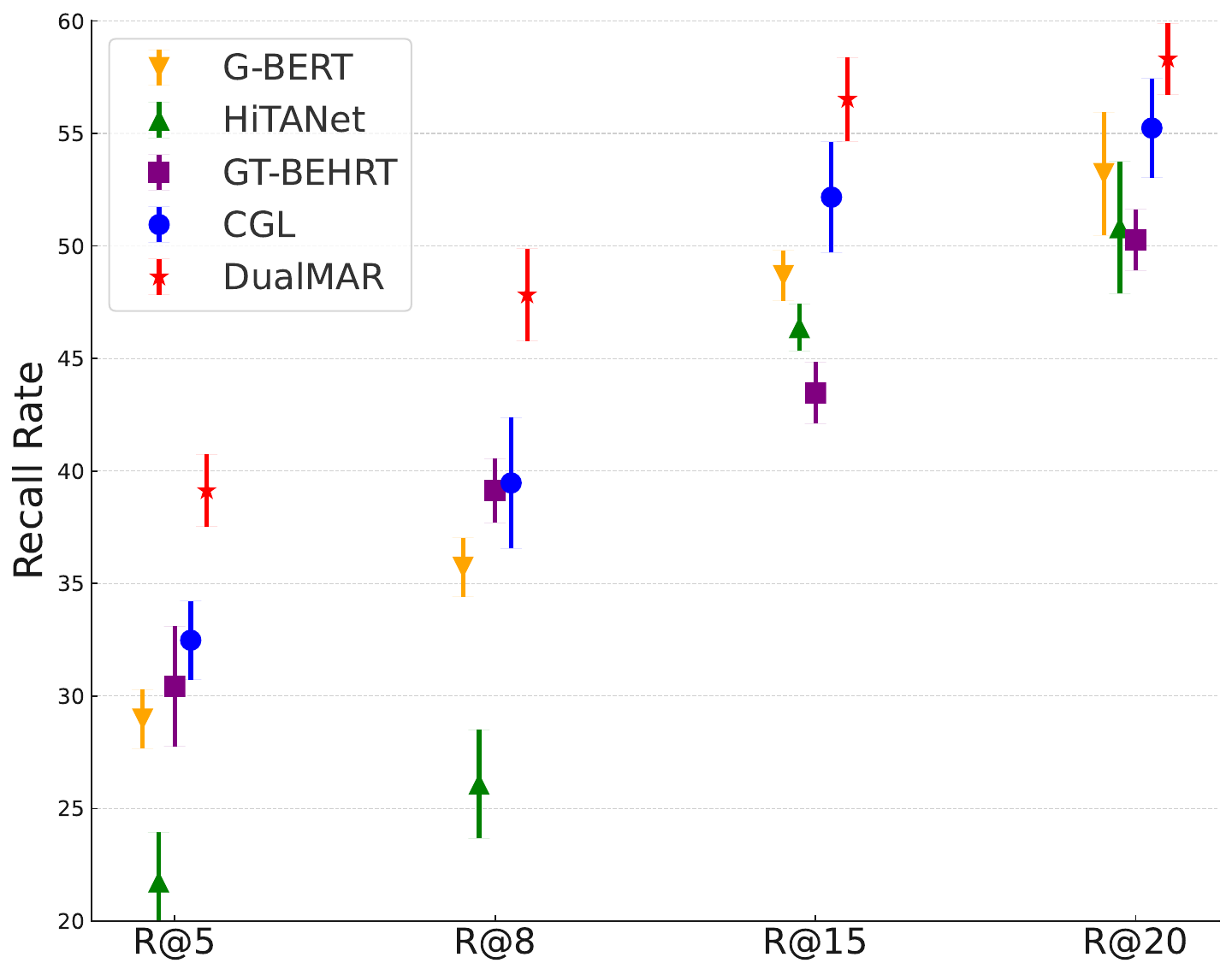}
        \caption{R@k in Less-frequent Diagnosis Prediction.}
        \label{fig:rare_diagnosis}
    \end{subfigure}
    \hfill
    \begin{subfigure}{0.48\linewidth}
        \centering
        \includegraphics[width=\linewidth]{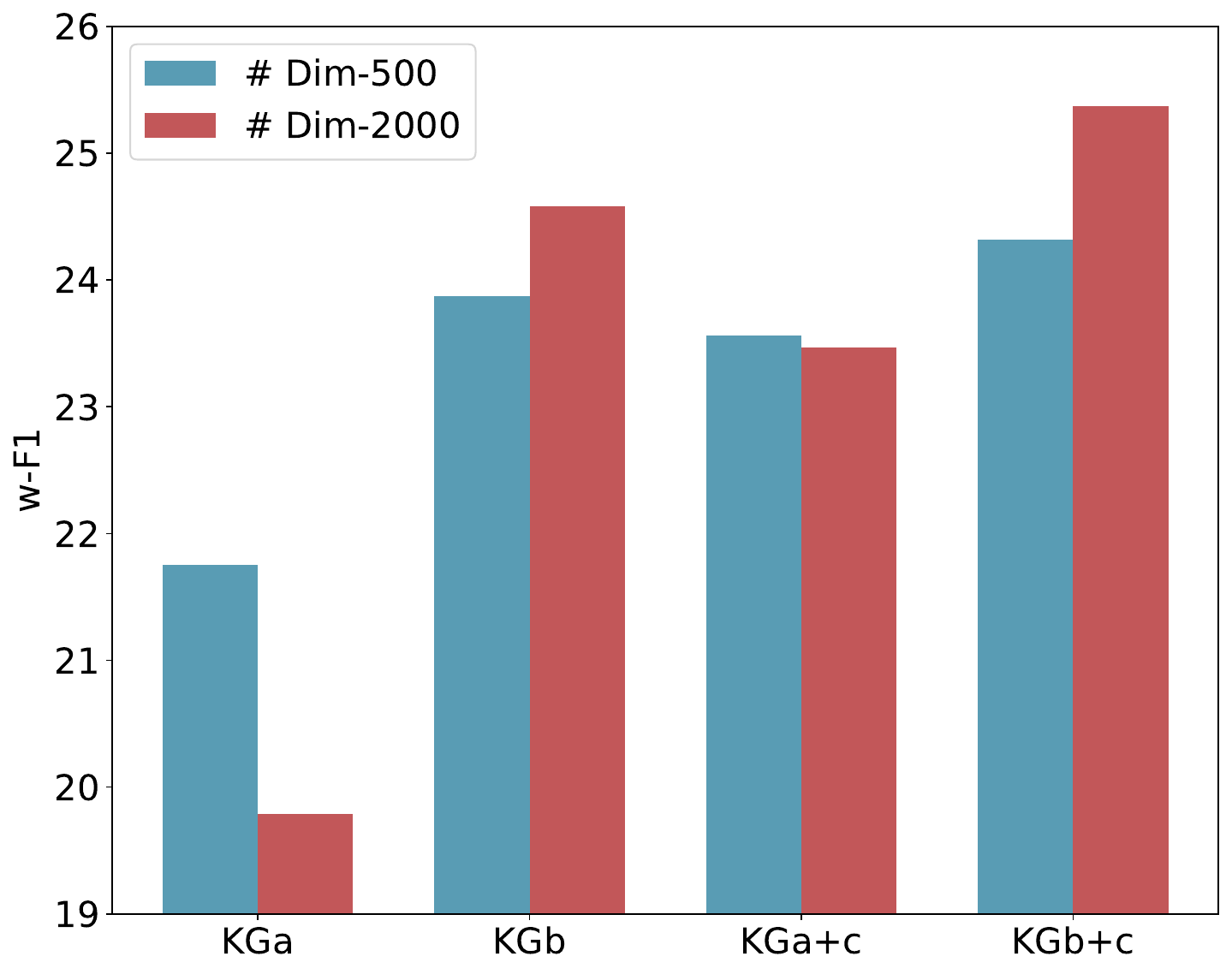}
        \caption{\textbf{w-$F_1$} of Diagnosis Prediction with different KGs.}
        \label{fig:kg_diagnosis}
    \end{subfigure}
    \caption{Results for the $1^{st}$ and $2^{nd}$ case studies, full results are shown in Appendix~\ref{app:detail_case}.}
    % \label{fig:combined_figure}
\end{figure}

To assess the importance of each component in our Diagnosis KG, we divided it into three parts: ICD9-Hierarchy \texttt{KG}\textsubscript{a}, Ontology-KG \texttt{KG}\textsubscript{b}, and LLM-KG \texttt{KG}\textsubscript{c}. We then evaluated the predictive performance of the model with these varying complexities.
% Table~\ref{tab:kg_embeddings}
Figure~\ref{fig:kg_diagnosis} shows how the complexity and embedding dimensions of KGs influence diagnosis prediction. Key observations include:
(1) LLM-generated triples (\texttt{KG}\textsubscript{c}) enhance comprehensive KG representation, as seen in \texttt{KG}\textsubscript{a+c} and \texttt{KG}\textsubscript{b+c}.
(2) Ontology-KG (\texttt{KG}\textsubscript{b}) from diverse data sources is more reliable for accurate predictions than ICD9-Hierarchy (\texttt{KG}\textsubscript{a}).
(3) Embedding dimensions significantly impact predictions; higher dimensions offer a deeper understanding but can lead to overfitting, especially with simpler KGs like \texttt{KG}\textsubscript{a}.

\subsubsection{Accessibility}
We evaluated the accessibility of \system{} with large-scale KGs, comparing it to GraphCare~\cite{jiang2023graphcare} using an NVIDIA RTX-4090 GPU with 24GB of memory.
For a fair comparison, we adapted \texttt{UMLS-GPT-KG} in GraphCare to our Diagnosis KG, adjusting only the batch size due to its end-to-end structure without pretraining module.
Table~\ref{tab:gpu_memory_usage} shows average GPU usage for diagnosis prediction.
GraphCare struggles with training on common commercial GPUs when transitioning from CCS KG to ICD-9 KG, even with small batch sizes, limiting its scalability with large KGs. 
In contrast, \system{} efficiently separates learning processes for KGs and EHR data, allowing it to scale to larger KGs and making it more deployable across various EHR tasks and datasets.

\setlength{\tabcolsep}{6pt}
\begin{table}[t]
    \centering
    \small
    \caption{Accessibility Analysis with Adjusted GraphCare.}
    \label{tab:gpu_memory_usage}
    \begin{tabular}{@{}lccc@{}}
        \toprule
        \textbf{Models} & \textbf{GPU Memory Usage} & \textbf{Runnable} \\
        \midrule
        \multicolumn{3}{@{}l}{\textbf{\system{}} - \textit{batch sizes: 256, 32, and 32}} \\
        \hspace{1em}- KG Embedding & 11.73 GB & \checkmark \\
        \hspace{1em}- Pretrain \& Finetune & 9.78 GB & \checkmark \\
        \hspace{1em}- Direct Training & 9.62 GB & \checkmark \\
        \midrule
        \multicolumn{3}{@{}l}{\textbf{GraphCare + Diagnosis KG}} \\
        \hspace{1em}- Batch Size: 2 & 33.95 GB & \texttimes \\
        \hspace{1em}- Batch Size: 4 & 42.57 GB & \texttimes \\
        \hspace{1em}- Batch Size: 8 & 98.23 GB & \texttimes \\
        \hspace{1em}- Batch Size: 16 & 172.04 GB & \texttimes \\
        \bottomrule
    \end{tabular}
\end{table}

\section{Conclusion}
We proposed \system{}, a dual-expertise framework for enhancing clinical prediction by jointly leveraging individual lab results and structured medical knowledge. 
\system{} generates a bi-hierarchical Diagnosis KG enriched via LLM-generated triples, capturing both semantic and hierarchical information essential for clinical prediction. 
Moreover, a lab-informed proxy-task learning aligns the model with real-world diagnostic processes by guiding stepwise clinical reasoning for diagnosis. 
These two perspectives are then integrated via a encoder-decoder architecture, enabling more precise and interpretable predictions. 
Experimental results demonstrate that \system{} consistently outperforms existing predictive baselines across multiple clinical tasks, which highlights the potential of combining structured knowledge and clinical evidence to advance EHR modeling.
% The results also offers a more faithful reflection of medical decision-making in data-driven systems. 
Overall, \system{} not only broadens the scope of medical-domain knowledge but also addresses the limitations of traditional EHR models by incorporating more complex and nuanced medical data. 
Future work could explore expanding \system{} to adapt to large-scale KGs and apply it to a broader range of healthcare prediction tasks beyond diagnostics. 
Additionally, this work highlights the crucial role of laboratory data in clinical decision-making, suggesting that further investigation into leveraging lab tests as auxiliary inputs for both diagnosis and treatment predictions remains a promising direction.

\appendices

\section*{Appendix}

\subsection{KG Embeddings Comparison}
\label{app:kge_compare}

\begin{table}[t]
    \centering
    \caption{Different Embedding Baselines Comparison (\%)}
    \label{tab:KGE}
    \begin{tabular}{lcccc}
    \toprule
    \textbf{MODEL} & \textbf{MRR} & \textbf{HITS@1} & \textbf{HITS@3} & \textbf{HITS@10} \\
    \midrule
    TransE & 54.77 & 45.90 & 58.30 & 72.40 \\
    RotatE & 57.82 & 47.95 & 62.56 & 75.28 \\
    ModE & 63.02 & 55.46 & 66.25 & 77.85 \\
    TripleRE & 64.02 & 57.21 & 68.03 & 79.32 \\
    HyperE & 59.03 & 53.69 & 64.82 & 74.93 \\
    \midrule
    HAKE & \textbf{67.37} & \textbf{60.23} & \textbf{70.62} & \textbf{81.21} \\
    \bottomrule
    \end{tabular}
\end{table}

Most existing hierarchical KG embedding methods leverage Euclidean or non-Euclidean geometries to encode hierarchy. 
Euclidean-based approaches typically use coordinate projections to distinguish entity levels and retrieve hierarchical information more efficiently than hyperbolic or spherical methods, which often require additional steps or higher computational cost. 
To further improve efficiency, we decouple embedding training from graph learning by using link prediction tasks, allowing efficient GPU memory usage even on complex KGs. In our approach, each entity is projected onto polar coordinates to facilitate efficient embedding retrieval. As shown in Table~\ref{tab:KGE}, our polar-space embedding method, adapted from HAKE~\cite{ZhangCZW20}, outperforms five widely-used baselines~\cite{BordesUGWY13, SunDNT19, NickelK17, FiondaP20} in capturing both hierarchical and semantic features, evaluated by MRR and HITS@k metrics (k = 1, 3, 10).

\subsection{The Significance of Improvement}
\label{app:detail_experiment}
To evaluate the statistical significance of performance differences between our proposed \system{} model and baseline models, we employed both the paired t-test and Wilcoxon signed-rank test. The t-test was chosen under the assumption of normally distributed performance metrics. To account for potential deviations from normality, the non-parametric Wilcoxon signed-rank test was also conducted. These tests allow us to robustly determine whether the observed improvements in \system{}'s performance are statistically significant compared to the baselines.

Here we only take the w-$F_1$ of Diagnosis Prediction in MIMIC-IV as an example to show the evaluation process on Table~\ref{tab:wilcoxon_ttest}. 
We observe that the Wilcoxon signed-rank test consistently yielded p-values of $1.95 \times 10^{-3}$, indicating significant differences between \system{} and all baseline models. The paired t-tests further confirmed these results, with all p-values well below $1 \times 10^{-10}$. The 95\% confidence intervals for the t-tests showed a clear positive difference, reinforcing the superior performance of \system{} in Diagnosis Prediction compared to each baseline. These results demonstrate the robustness of \system{}’s improvements in weighted $F_1$ scores over the baselines.

\setlength{\tabcolsep}{5pt}
\begin{table}[t]
    \centering
    \small
    \caption{Wilcoxon Signed-Rank Test and t-Test Results for \system{} vs. Baseline Models with 10 runs. ``p'' means p-value for certain evaluation method, and ``CI'' means Confidence Interval calculated by T-testing method.}
    \label{tab:wilcoxon_ttest}
    \begin{tabular}{@{}lccc@{}}
        \toprule
        \textbf{Models} & \textbf{Wilcoxon p} & \textbf{T-test p} & \textbf{95\% CI} \\
        \hline
        \system{} vs. RETAIN & 1.95e-03 & 1.62e-14 & (7.41, 7.80) \\
        \system{} vs. Timeline & 1.95e-03 & 2.01e-14 & (7.17, 7.53) \\
        \system{} vs. GCT & 1.95e-03 & 6.27e-10 & (6.46, 7.39) \\
        \system{} vs. CGL & 1.95e-03 & 9.77e-13 & (5.93, 6.39) \\
        \system{} vs. SeqCare & 1.95e-03 & 5.53e-13 & (5.31, 5.67) \\
        \system{} vs. RAM-EHR & 1.95e-03 & 1.23e-12 & (4.87, 5.21) \\
        \system{} vs. G-BERT & 1.95e-03 & 3.46e-11 & (4.49, 4.88) \\
        \system{} vs. HiTANet & 1.95e-03 & 7.65e-13 & (5.62, 6.01) \\
        \system{} vs. Med-BERT & 1.95e-03 & 1.99e-12 & (5.89, 6.27) \\
        \system{} vs. GT-BEHRT & 1.95e-03 & 4.57e-11 & (3.98, 4.32) \\
        \bottomrule
    \end{tabular}
\end{table}

\begin{table}[t]
    \centering
    \small
    \caption{Less Frequent Codes Diagnosis Prediction.}
    \label{tab:rare_diagnosis}
    \setlength{\tabcolsep}{5pt}
    \begin{tabular}{lcccccc}
        \toprule
         & \multicolumn{3}{c}{\textbf{Rare-10}} & \multicolumn{3}{c}{\textbf{Rare-20}} \\
        \textbf{Model} & \textbf{R@5} & \textbf{R@8} & \textbf{R@15} & \textbf{R@5} & \textbf{R@8} & \textbf{R@15} \\
        \midrule
        CGL & 32.48 & 39.47 & 52.17 & 12.53 & 16.03 & 24.40 \\
        G-BERT & 28.97 & 35.72 & 48.68 & \textbf{14.95} & 17.11 & 25.10 \\
        HiTANet & 21.74 & 26.09 & 46.38 & 11.99 & 15.43 & 22.49 \\
        GT-BEHRT & 30.43 & 39.13 & 43.48 & 12.95 & 17.56 & 23.79 \\
        \system{} & \textbf{39.13} & \textbf{47.83} & \textbf{56.52} & 14.28 & \textbf{18.00} & \textbf{26.30} \\
        \bottomrule
    \end{tabular}
\end{table}

\begin{table}[t]
    \centering
    \small
    \caption{Diagnosis Prediction Results by Different KGs.}
    \label{tab:kg_embeddings}
    \setlength{\tabcolsep}{2pt}
    \begin{tabular}{@{}llccc@{}}
        \toprule
        \textbf{Models} & \textbf{\# Dims} & \textbf{w-$F_1$} & \textbf{R@10} & \textbf{R@20} \\
        \midrule
        \system{}-\texttt{KG}$_{a}$   & 500  & 21.75$_{(0.3)}$ & 33.27$_{(0.4)}$ & 35.32$_{(0.4)}$ \\
        \system{}-\texttt{KG}$_{a}$   & 2000 & 19.79$_{(0.5)}$ & 32.12$_{(0.4)}$ & 33.57$_{(0.4)}$ \\
        \system{}-\texttt{KG}$_{b}$   & 500  & 23.87$_{(0.3)}$ & 37.73$_{(0.4)}$ & 39.18$_{(0.4)}$ \\
        \system{}-\texttt{KG}$_{b}$   & 2000 & 24.58$_{(0.3)}$ & 39.27$_{(0.4)}$ & 40.32$_{(0.4)}$ \\
        \system{}-\texttt{KG}$_{a+c}$ & 500  & 23.56$_{(0.2)}$ & 37.43$_{(0.3)}$ & 38.74$_{(0.4)}$ \\
        \system{}-\texttt{KG}$_{a+c}$ & 2000 & 23.47$_{(0.3)}$ & 37.89$_{(0.4)}$ & 38.54$_{(0.4)}$ \\
        \system{}-\texttt{KG}$_{b+c}$ & 500  & 24.32$_{(0.3)}$ & 39.73$_{(0.4)}$ & 40.15$_{(0.4)}$ \\
        \system{}-\texttt{KG}$_{b+c}$ & 2000 & \textbf{25.37}$_{(\textbf{0.3})}$ & \textbf{40.52}$_{(\textbf{0.2})}$ & \textbf{41.86}$_{(\textbf{0.3})}$ \\
        \bottomrule
    \end{tabular}
\end{table}

\subsection{Details in Case Studies}
\label{app:detail_case}

We design experiments focusing on predicting less frequent condition codes in MIMIC-III, similar to rare ICD code prediction tasks. 
We conduct two subtasks:
\begin{itemize}
    \item \texttt{Rare-20}: Predicting 200 ICD codes that occur fewer than 20 times in MIMIC-III.
    \item \texttt{Rare-10}: Predicting 100 ICD codes that occur fewer than 10 times in MIMIC-III.
\end{itemize}

Table~\ref{tab:rare_diagnosis} shows the full results of both \texttt{Rare-10} and \texttt{Rare-20}.
It highlights \system{}'s superior performance in predicting less frequent ICD codes across both subtasks, \texttt{Rare-10} and \texttt{Rare-20}. Specifically, \system{} consistently outperforms baseline models, achieving the highest Recall at all thresholds (R@5, R@8, R@15). Notably, for \texttt{Rare-20}, \system{} significantly surpasses other models, particularly in R@8 and R@15, reflecting its strong capability in handling rare condition codes. These results underscore the effectiveness of \system{} in scenarios where capturing rare but critical diagnoses is crucial.

Table~\ref{tab:kg_embeddings} shows the full results of \system{} with varying KGs and embedding sizes.
Beyond analysis for w-$F_1$ in the case study, full table also demonstrates the Recall performance of \system{}. The results indicate that incorporating KG components (\texttt{KG}\textsubscript{b+c}) consistently yields the highest Recall across R@10 and R@20, particularly with 2000-dimensional embeddings. Notably, \texttt{KG}\textsubscript{b} also shows strong performance, suggesting the significant contributions of fundamental Ontology-KG in enhancing diagnostic predictions. The comparison highlights the critical role of augmenting KG to achieve superior recall in diagnosis prediction tasks.

\subsection{Prompting LLM}
\label{app:prompt_llm}
The following template in Table~\ref{tab:prompt} is used to instruct LLMs to convert textual information into structured triples.

\begin{table*}[p]
\centering
\small
\caption{General Prompting Framework for Triple Generation Tasks of LLMs}
\label{tab:prompt}
\begin{tabular}{|l|p{15cm}|}
\hline
\textbf{Name} & \textbf{Prompt Template} \\ 
\hline

\textbf{Variables} \quad & 
\begin{minipage}[t]{\linewidth}
\raggedright 
\texttt{<category>}: The prompt is not only suitable for disease but also available for concepts within medical domains such as medication and treatment. In this study, only condition concepts are considered.\\
\texttt{<term>}: The concept name. In this study, it means disease names provided by web-scraped text.\\
\texttt{<topics>}: The topics related to crawled text, and it has been provided when we crawled them from websites.\\
\texttt{<text>}: The content of crawled text. \\
\quad \\

\end{minipage}\\

\hline

\textbf{Skeleton} & 
\begin{minipage}[t]{\linewidth}
\raggedright 
Given a crawled text about specific topic of certain \texttt{<category>}, please find triples related to the given \texttt{<category>} in terms of crawled text.
\begin{itemize}
    \item Filling triples in updates based on given information and strictly following output style of example updates.
    \item Each update should follow the format of [ENTITY 1, RELATIONSHIP, ENTITY 2] with directed edge.
    \item Both ENTITY 1 and ENTITY 2 should be noun, and one of them must be \texttt{<term>}.
    \item Just output each unique triple once, don’t output repeatedly.
    \item It is possible that \texttt{<category>} name not exactly matched in crawled text (abbreviated or partly matched), consider it as the same thing.
\end{itemize}

\quad \\
Example: \\
\#\# An example demo is shown below...\\
\quad \\

Given a paragraph about specific topic of certain \texttt{<category>}, please find triples related to the given \texttt{<category>} in the text. \\
\quad \\
\textbf{Given Information:} \\
\texttt{<category>} \textbf{Name}: \texttt{<term>} \\
\textbf{Topics}: \texttt{<topics>} \\
\textbf{Text}: \texttt{<text>} \\
\quad\\
\textbf{Updates:} \\
\quad\\
\#\# LLM's output is expected to generate here in terms of given information...\\
\#\# \text{[Head Entity, Relation, Tail Entity]}...\\
\quad\\
\end{minipage} \\ 

\hline

\textbf{Example} \qquad & 
\begin{minipage}[t]{\linewidth}
\raggedright
\textbf{Disease Name:} Heart Failure \\
\textbf{Topics:} Overview \\
\textbf{Text:} \\
\textit{Heart failure occurs when the heart muscle doesn’t pump blood as well as it should. When this happens, blood often backs up and fluid can build up in the lungs, causing shortness of breath. Certain heart conditions gradually leave the heart too weak or stiff to fill and pump blood properly. These conditions include narrowed arteries in the heart and high blood pressure. Proper treatment may improve the symptoms of heart failure and may help some people live longer. Lifestyle changes can improve quality of life. Try to lose weight, exercise, use less salt and manage stress. But heart failure can be life-threatening. People with heart failure may have severe symptoms. Some may need a heart transplant or a device to help the heart pump blood. Heart failure is sometimes called congestive heart failure.} \\
\quad\\
\textbf{Updates:}
\begin{itemize}
    \item \text{[Heart Failure, IS\_CAUSED\_BY, Narrowed Arteries]},
    \item \text{[Heart Failure, IS\_CAUSED\_BY, High Blood Pressure]},
    \item \text{[Heart Failure, HAS\_SYMPTOMS, Shortness of Breath]},
    \item \text{[Heart Failure, HAS\_SYMPTOMS, Fluid Build-up in Lungs]},
    \item \text{[Heart Failure, NEEDS\_TREATMENT, Proper Treatment]},
    \item \text{[Heart Failure, NEEDS\_TREATMENT, Lifestyle Changes]}
\end{itemize}

\quad \\
\#\# You can use more examples to refine output of LLMs, but only one example is also fine for this task. \\
\quad \\
\end{minipage} \\ \hline
\end{tabular}
\end{table*}

\section*{Acknowledgment}

This work is supported in part by the US National Science Foundation under grants 2047843 and 2437621. Any opinions, findings, and conclusions or recommendations expressed in this material are those of the authors and do not necessarily reflect the views of the National Science Foundation.

\section*{References}
\bibliographystyle{IEEEtran}
\bibliography{main}

\begin{thebibliography}{10}
\providecommand{\url}[1]{#1}
\csname url@samestyle\endcsname
\providecommand{\newblock}{\relax}
\providecommand{\bibinfo}[2]{#2}
\providecommand{\BIBentrySTDinterwordspacing}{\spaceskip=0pt\relax}
\providecommand{\BIBentryALTinterwordstretchfactor}{4}
\providecommand{\BIBentryALTinterwordspacing}{\spaceskip=\fontdimen2\font plus
\BIBentryALTinterwordstretchfactor\fontdimen3\font minus \fontdimen4\font\relax}
\providecommand{\BIBforeignlanguage}[2]{{%
\expandafter\ifx\csname l@#1\endcsname\relax
\typeout{** WARNING: IEEEtran.bst: No hyphenation pattern has been}%
\typeout{** loaded for the language `#1'. Using the pattern for}%
\typeout{** the default language instead.}%
\else
\language=\csname l@#1\endcsname
\fi
#2}}
\providecommand{\BIBdecl}{\relax}
\BIBdecl

\bibitem{LuRCKN21}
C.~Lu, C.~K. Reddy, P.~Chakraborty, S.~Kleinberg, and Y.~Ning, ``Collaborative graph learning with auxiliary text for temporal event prediction in healthcare,'' in \emph{Proceedings of the Thirtieth International Joint Conference on Artificial Intelligence, {IJCAI} 2021}, 2021, pp. 3529--3535.

\bibitem{ChoiXLDFXD20}
E.~Choi, Z.~Xu, Y.~Li, M.~Dusenberry, G.~Flores, E.~Xue, and A.~M. Dai, ``Learning the graphical structure of electronic health records with graph convolutional transformer,'' in \emph{The Thirty-Fourth {AAAI} Conference on Artificial Intelligence, {AAAI} 2020}, 2020, pp. 606--613.

\bibitem{shickel2017deep}
B.~Shickel, P.~J. Tighe, A.~Bihorac, and P.~Rashidi, ``Deep ehr: a survey of recent advances in deep learning techniques for electronic health record (ehr) analysis,'' \emph{IEEE journal of biomedical and health informatics}, vol.~22, no.~5, pp. 1589--1604, 2017.

\bibitem{ShangMXS19}
J.~Shang, T.~Ma, C.~Xiao, and J.~Sun, ``Pre-training of graph augmented transformers for medication recommendation,'' in \emph{Proceedings of the Twenty-Eighth International Joint Conference on Artificial Intelligence, {IJCAI} 2019}, 2019, pp. 5953--5959.

\bibitem{LuRN23}
C.~Lu, C.~K. Reddy, and Y.~Ning, ``Self-supervised graph learning with hyperbolic embedding for temporal health event prediction,'' \emph{{IEEE} Trans. Cybern.}, vol.~53, no.~4, pp. 2124--2136, 2023.

\bibitem{poulain2024graph}
R.~Poulain and R.~Beheshti, ``Graph transformers on {EHR}s: Better representation improves downstream performance,'' in \emph{The Twelfth International Conference on Learning Representations}, 2024.

\bibitem{ChoiBSSS17}
E.~Choi, M.~T. Bahadori, L.~Song, W.~F. Stewart, and J.~Sun, ``{GRAM:} graph-based attention model for healthcare representation learning,'' in \emph{Proceedings of the 23rd {ACM} {SIGKDD}, 2017}, 2017, pp. 787--795.

\bibitem{MaYXCZG18}
F.~Ma, Q.~You, H.~Xiao, R.~Chitta, J.~Zhou, and J.~Gao, ``{KAME:} knowledge-based attention model for diagnosis prediction in healthcare,'' in \emph{Proceedings of the 27th {ACM} International Conference on Information and Knowledge Management, {CIKM} 2018, Torino, Italy, October 22-26, 2018}, 2018, pp. 743--752.

\bibitem{jiang2023graphcare}
P.~Jiang, C.~Xiao, A.~R. Cross, and J.~Sun, ``Graphcare: Enhancing healthcare predictions with personalized knowledge graphs,'' in \emph{The Twelfth International Conference on Learning Representations}, 2023.

\bibitem{world1988international}
W.~H. Organization \emph{et~al.}, ``International classification of diseases—ninth revision (icd-9),'' \emph{Weekly Epidemiological Record= Relev{\'e} {\'e}pid{\'e}miologique hebdomadaire}, vol.~63, no.~45, pp. 343--344, 1988.

\bibitem{xu2023seqcare}
Y.~Xu, X.~Chu, K.~Yang, Z.~Wang, P.~Zou, H.~Ding, J.~Zhao, Y.~Wang, and B.~Xie, ``Seqcare: Sequential training with external medical knowledge graph for diagnosis prediction in healthcare data,'' in \emph{Proceedings of the ACM Web Conference 2023}, 2023, pp. 2819--2830.

\bibitem{xu2024ram}
R.~Xu, W.~Shi, Y.~Yu, Y.~Zhuang, B.~Jin, M.~D. Wang, J.~C. Ho, and C.~Yang, ``Ram-ehr: Retrieval augmentation meets clinical predictions on electronic health records,'' in \emph{Proceedings of the 62nd Annual Meeting of the Association for Computational Linguistics}, 2024.

\bibitem{Rasmy0XTZ21}
L.~Rasmy, Y.~Xiang, Z.~Xie, C.~Tao, and D.~Zhi, ``Med-bert: pretrained contextualized embeddings on large-scale structured electronic health records for disease prediction,'' \emph{npj Digit. Medicine}, vol.~4, 2021.

\bibitem{PrakashCRV21}
P.~K.~S. Prakash, S.~Chilukuri, N.~Ranade, and S.~Viswanathan, ``Rarebert: Transformer architecture for rare disease patient identification using administrative claims,'' in \emph{Thirty-Fifth {AAAI} Conference on Artificial Intelligence, {AAAI} 2021}, 2021, pp. 453--460.

\bibitem{boll2024graph}
H.~O. Boll, A.~Amirahmadi, M.~M. Ghazani, W.~O. de~Morais, E.~P. de~Freitas, A.~Soliman, F.~Etminani, S.~Byttner, and M.~Recamonde-Mendoza, ``Graph neural networks for clinical risk prediction based on electronic health records: A survey.'' \emph{J. Biomed. Informatics}, vol. 151, p. 104616, 2024.

\bibitem{sutton2020overview}
R.~T. Sutton, D.~Pincock, D.~C. Baumgart, D.~C. Sadowski, R.~N. Fedorak, and K.~I. Kroeker, ``An overview of clinical decision support systems: benefits, risks, and strategies for success,'' \emph{NPJ digital medicine}, vol.~3, no.~1, p.~17, 2020.

\bibitem{ChoiBSKSS16}
E.~Choi, M.~T. Bahadori, J.~Sun, J.~Kulas, A.~Schuetz, and W.~F. Stewart, ``{RETAIN:} an interpretable predictive model for healthcare using reverse time attention mechanism,'' in \emph{Advances in Neural Information Processing Systems 29: Annual Conference on Neural Information Processing Systems 2016}, 2016, pp. 3504--3512.

\bibitem{ChoiBSSS16}
E.~Choi, M.~T. Bahadori, A.~Schuetz, W.~F. Stewart, and J.~Sun, ``Doctor {AI:} predicting clinical events via recurrent neural networks,'' in \emph{Proceedings of the 1st Machine Learning in Health Care}, vol.~56, 2016, pp. 301--318.

\bibitem{MaCZYSG17}
F.~Ma, R.~Chitta, J.~Zhou, Q.~You, T.~Sun, and J.~Gao, ``Dipole: Diagnosis prediction in healthcare via attention-based bidirectional recurrent neural networks,'' in \emph{Proceedings of the 23rd {ACM} {SIGKDD} International Conference on Knowledge Discovery and Data Mining, Halifax, NS, Canada, August 13 - 17, 2017}, 2017, pp. 1903--1911.

\bibitem{BaiZEV18}
T.~Bai, S.~Zhang, B.~L. Egleston, and S.~Vucetic, ``Interpretable representation learning for healthcare via capturing disease progression through time,'' in \emph{Proceedings of the 24th {ACM} {SIGKDD} International Conference on Knowledge Discovery {\&} Data Mining, {KDD} 2018}, 2018, pp. 43--51.

\bibitem{MaZWRWTMGG20}
L.~Ma, C.~Zhang, Y.~Wang, W.~Ruan, J.~Wang, W.~Tang, X.~Ma, X.~Gao, and J.~Gao, ``Concare: Personalized clinical feature embedding via capturing the healthcare context,'' in \emph{The Thirty-Fourth {AAAI} Conference on Artificial Intelligence, {AAAI} 2020}, 2020, pp. 833--840.

\bibitem{NguyenTWV17}
P.~Nguyen, T.~Tran, N.~Wickramasinghe, and S.~Venkatesh, ``\texttt{Deepr}: {A} convolutional net for medical records,'' \emph{{IEEE} J. Biomed. Health Informatics}, vol.~21, no.~1, pp. 22--30, 2017.

\bibitem{MaGWZWRTGM20}
L.~Ma, J.~Gao, Y.~Wang, C.~Zhang, J.~Wang, W.~Ruan, W.~Tang, X.~Gao, and X.~Ma, ``Adacare: Explainable clinical health status representation learning via scale-adaptive feature extraction and recalibration,'' in \emph{The Thirty-Fourth {AAAI} Conference on Artificial Intelligence, {AAAI} 2020}, 2020, pp. 825--832.

\bibitem{rasmy2021med}
L.~Rasmy, Y.~Xiang, Z.~Xie, C.~Tao, and D.~Zhi, ``Med-bert: pretrained contextualized embeddings on large-scale structured electronic health records for disease prediction,'' \emph{NPJ digital medicine}, vol.~4, no.~1, p.~86, 2021.

\bibitem{ChoiXSS18}
E.~Choi, C.~Xiao, W.~F. Stewart, and J.~Sun, ``Mime: Multilevel medical embedding of electronic health records for predictive healthcare,'' in \emph{Neural Information Processing Systems 2018, NeurIPS 2018}, 2018, pp. 4552--4562.

\bibitem{ParkBKKC22}
S.~Park, S.~Bae, J.~Kim, T.~Kim, and E.~Choi, ``Graph-text multi-modal pre-training for medical representation learning,'' in \emph{Conference on Health, Inference, and Learning, {CHIL} 2022, 7-8 April 2022, Virtual Event}, vol. 174, 2022, pp. 261--281.

\bibitem{zou2024ai}
X.~Zou, W.~He, Y.~Huang, Y.~Ouyang, Z.~Zhang, Y.~Wu, Y.~Wu, L.~Feng, S.~Wu, M.~Yang \emph{et~al.}, ``Ai-driven diagnostic assistance in medical inquiry: Reinforcement learning algorithm development and validation,'' \emph{Journal of Medical Internet Research}, vol.~26, p. e54616, 2024.

\bibitem{smith2023bias}
B.~Smith, A.~Khojandi, and R.~Vasudevan, ``Bias in reinforcement learning: A review in healthcare applications,'' \emph{ACM Computing Surveys}, vol.~56, no.~2, pp. 1--17, 2023.

\bibitem{nguyen2024carer}
T.~Nguyen, T.~Huynh, M.~H. Phan, Q.~V.~H. Nguyen, and P.~Le~Nguyen, ``Carer-clinical reasoning-enhanced representation for temporal health risk prediction,'' in \emph{Proceedings of the 2024 Conference on Empirical Methods in Natural Language Processing}, 2024, pp. 10\,392--10\,407.

\bibitem{jiang2025reasoningenhanced}
\BIBentryALTinterwordspacing
P.~Jiang, C.~Xiao, M.~Jiang, P.~Bhatia, T.~Kass-Hout, J.~Sun, and J.~Han, ``Reasoning-enhanced healthcare predictions with knowledge graph community retrieval,'' in \emph{The Thirteenth International Conference on Learning Representations}, 2025. [Online]. Available: \url{https://openreview.net/forum?id=8fLgt7PQza}
\BIBentrySTDinterwordspacing

\bibitem{cuietal2023}
H.~Cui, J.~Lu, S.~Wang, R.~Xu, W.~Ma, S.~Yu, Y.~Yu, X.~Kan, C.~Ling, J.~C. Ho, F.~Wang, and C.~Yang, ``A survey on knowledge graphs for healthcare: Resources, applications, and promises,'' \emph{CoRR}, vol. abs/2306.04802, 2023.

\bibitem{bodenreider2004unified}
O.~Bodenreider, ``The unified medical language system (umls): integrating biomedical terminology,'' \emph{Nucleic acids research}, vol.~32, no. suppl\_1, pp. D267--D270, 2004.

\bibitem{knox2024drugbank}
C.~Knox, M.~Wilson, C.~M. Klinger, M.~Franklin, E.~Oler, A.~Wilson, A.~Pon, J.~Cox, N.~E. Chin, S.~A. Strawbridge \emph{et~al.}, ``Drugbank 6.0: the drugbank knowledgebase for 2024,'' \emph{Nucleic acids research}, vol.~52, no.~D1, pp. D1265--D1275, 2024.

\bibitem{ursu2016drugcentral}
O.~Ursu, J.~Holmes, J.~Knockel, C.~G. Bologa, J.~J. Yang, S.~L. Mathias, S.~J. Nelson, and T.~I. Oprea, ``Drugcentral: online drug compendium,'' \emph{Nucleic acids research}, p. gkw993, 2016.

\bibitem{gargano2024mode}
M.~A. Gargano, N.~Matentzoglu, B.~Coleman, E.~B. Addo-Lartey, A.~V. Anagnostopoulos, J.~Anderton, P.~Avillach, A.~M. Bagley, E.~Bak{\v{s}}tein, J.~P. Balhoff \emph{et~al.}, ``The human phenotype ontology in 2024: phenotypes around the world,'' \emph{Nucleic acids research}, vol.~52, no.~D1, pp. D1333--D1346, 2024.

\bibitem{kuhn2016sider}
M.~Kuhn, I.~Letunic, L.~J. Jensen, and P.~Bork, ``The sider database of drugs and side effects,'' \emph{Nucleic acids research}, vol.~44, no.~D1, pp. D1075--D1079, 2016.

\bibitem{brennan1998mayo}
M.~D. Brennan, K.~M. Miner, and R.~A. Rizza, ``The mayo clinic,'' \emph{The Journal of Clinical Endocrinology \& Metabolism}, vol.~83, no.~10, pp. 3427--3434, 1998.

\bibitem{weinreich2008orphanet}
S.~S. Weinreich, R.~Mangon, J.~Sikkens, M.~E. Teeuw, and M.~Cornel, ``Orphanet: a european database for rare diseases,'' \emph{Nederlands tijdschrift voor geneeskunde}, vol. 152, no.~9, pp. 518--519, 2008.

\bibitem{schieppati2008rare}
A.~Schieppati, J.-I. Henter, E.~Daina, and A.~Aperia, ``Why rare diseases are an important medical and social issue,'' \emph{The Lancet}, vol. 371, no. 9629, pp. 2039--2041, 2008.

\bibitem{grattafiori2024llama}
A.~Grattafiori, A.~Dubey, A.~Jauhri, A.~Pandey, A.~Kadian, A.~Al-Dahle, A.~Letman, A.~Mathur, A.~Schelten, A.~Vaughan \emph{et~al.}, ``The llama 3 herd of models,'' \emph{arXiv preprint arXiv:2407.21783}, 2024.

\bibitem{guo2025deepseek}
D.~Guo, D.~Yang, H.~Zhang, J.~Song, R.~Zhang, R.~Xu, Q.~Zhu, S.~Ma, P.~Wang, X.~Bi \emph{et~al.}, ``Deepseek-r1: Incentivizing reasoning capability in llms via reinforcement learning,'' \emph{arXiv preprint arXiv:2501.12948}, 2025.

\bibitem{nahler2009anatomical}
G.~Nahler and G.~Nahler, ``Anatomical therapeutic chemical classification system (atc),'' \emph{Dictionary of pharmaceutical medicine}, pp. 8--8, 2009.

\bibitem{chandak2023building}
P.~Chandak, K.~Huang, and M.~Zitnik, ``Building a knowledge graph to enable precision medicine,'' \emph{Scientific Data}, vol.~10, no.~1, p.~67, 2023.

\bibitem{ZhangCZW20}
Z.~Zhang, J.~Cai, Y.~Zhang, and J.~Wang, ``Learning hierarchy-aware knowledge graph embeddings for link prediction,'' in \emph{The Thirty-Fourth {AAAI} Conference on Artificial Intelligence, {AAAI} 2020, The Thirty-Second Innovative Applications of Artificial Intelligence Conference, {IAAI} 2020, The Tenth {AAAI} Symposium on Educational Advances in Artificial Intelligence, {EAAI} 2020, New York, NY, USA, February 7-12, 2020}, 2020, pp. 3065--3072.

\bibitem{BordesUGWY13}
A.~Bordes, N.~Usunier, A.~Garc{\'{\i}}a{-}Dur{\'{a}}n, J.~Weston, and O.~Yakhnenko, ``Translating embeddings for modeling multi-relational data,'' in \emph{Advances in Neural Information Processing Systems 26: 27th Annual Conference on Neural Information Processing Systems 2013}, 2013, pp. 2787--2795.

\bibitem{NickelK17}
M.~Nickel and D.~Kiela, ``Poincar{\'{e}} embeddings for learning hierarchical representations,'' in \emph{Advances in Neural Information Processing Systems 30: Annual Conference on Neural Information Processing Systems 2017, December 4-9, 2017, Long Beach, CA, {USA}}, 2017, pp. 6338--6347.

\bibitem{SunDNT19}
Z.~Sun, Z.~Deng, J.~Nie, and J.~Tang, ``Rotate: Knowledge graph embedding by relational rotation in complex space,'' in \emph{7th International Conference on Learning Representations, {ICLR} 2019, New Orleans, LA, USA, May 6-9, 2019}, 2019.

\bibitem{FiondaP20}
V.~Fionda and G.~Pirr{\`{o}}, ``Learning triple embeddings from knowledge graphs,'' in \emph{The Thirty-Fourth {AAAI} Conference on Artificial Intelligence, {AAAI} 2020, The Thirty-Second Innovative Applications of Artificial Intelligence Conference, {IAAI} 2020, The Tenth {AAAI} Symposium on Educational Advances in Artificial Intelligence, {EAAI} 2020, New York, NY, USA, February 7-12, 2020}.\hskip 1em plus 0.5em minus 0.4em\relax {AAAI} Press, 2020, pp. 3874--3881.

\bibitem{johnson2016mimic}
A.~E. Johnson, T.~J. Pollard, L.~Shen, L.-w.~H. Lehman, M.~Feng, M.~Ghassemi, B.~Moody, P.~Szolovits, L.~Anthony~Celi, and R.~G. Mark, ``Mimic-iii, a freely accessible critical care database,'' \emph{Scientific data}, vol.~3, no.~1, pp. 1--9, 2016.

\bibitem{johnson2020mimic}
A.~Johnson, L.~Bulgarelli, T.~Pollard, S.~Horng, L.~A. Celi, and R.~Mark, ``Mimic-iv,'' \emph{PhysioNet. Available online at: https://physionet. org/content/mimiciv/1.0/(accessed August 23, 2021)}, pp. 49--55, 2020.

\bibitem{LuoYXM20}
J.~Luo, M.~Ye, C.~Xiao, and F.~Ma, ``Hitanet: Hierarchical time-aware attention networks for risk prediction on electronic health records,'' in \emph{{KDD} '20: The 26th {ACM} {SIGKDD} Conference on Knowledge Discovery and Data Mining, Virtual Event, CA, USA, August 23-27, 2020}, 2020, pp. 647--656.

\bibitem{embedding2024}
\BIBentryALTinterwordspacing
{OpenAI}, ``Embeddings,'' 2024. [Online]. Available: \url{https://platform.openai.com/docs/guides/embeddings}
\BIBentrySTDinterwordspacing

\end{thebibliography}

\end{document}